# GazeRIB: Systematic Evaluation of Multi-Reader Gaze-Guided Chest X-ray Learning

Nattakorn Kittisut[1], Roy M. Gabriel[1], Marly van Assen[2], Mohammadreza Zandehshahvar[1], Carlo N. De Cecco[2], Elizabeth A. Krupinski[2], Ali Adibi[1*]

1School of Electrical and Computer Engineering, Georgia Institute of Technology, Atlanta, GA 30332, USA.

2Department of Radiology and Imaging Sciences, Emory University, Atlanta, GA 30307, USA.

*Corresponding author: ali.adibi@ece.gatech.edu

**Abstract**

Radiologists' visual attention can supervise medical image learning, but how reader panel composition affects its value remains untested. We evaluated GazeRIB (Gaze-guided Radiologist-Informed Blackout), which masks unviewed regions during training and uses unmasked images at inference. Eight experts repeatedly read the same 80 chest X-rays (CXRs). We compared individual readers with three-, five- and eight-reader panels across seven architectures and 10 seeds, evaluating 910 models on a separate 2,897-CXR cohort for four-class severity prediction. Every panel improved macro F1 over the no-gaze baseline by 5.33-7.47 percentage points (pp; Bonferroni-adjusted $p_{\text{adj}} < 0.007$); no individual reader reached significance on this endpoint. Three readers captured 73-80% of the eight-reader gain in macro F1, the area under the receiver-operating-characteristic curve (ROC-AUC) and the area under the precision-recall curve (AU-PRC). On external validation without retraining, all four panels improved ROC-AUC on SIIM-FISABIO-RSNA by +1.93 to +4.24 pp ($p \leq 0.004$); on MIMIC-Eye no panel differed significantly from baseline, whereas the single-reader average declined (-1.40 pp, $p = 0.008$). Reader composition therefore influenced performance, although transfer depended on the cohort.

## 1 Introduction

Chest radiography remains the first-line imaging test for acute lung disease, and the severity a radiologist assigns directly informs triage, escalation and treatment decisions. Deep learning (DL) has transformed medical imaging, especially chest X-ray (CXR) analysis, with convolutional neural networks (CNNs) and transformer models approaching radiologist-level performance on disease detection and severity scoring [1-5]. Gains are driven by high-capacity architectures, curated medical datasets, and transfer learning from ImageNet. In practice, models are fine-tuned on small labeled CXR cohorts for domain-specific tasks such as lung disease severity classification. However, such models may fail to attend to clinically meaningful regions, limiting generalization [6-8].

Eye-tracking (ET) provides a direct window into radiologists' visual search and has emerged as a complementary signal for understanding and improving CXR classification. Kundel et al. showed that fixation patterns during CXR reading reveal both search and recognition errors in lung nodule detection [9], and subsequent work established ET as a tool for characterizing diagnostic workflows and guiding artificial intelligence (AI) development [10, 11]. Recent multimodal CXR datasets combine images, reports, and eye-gaze [12-14], enabling AI systems that learn from both labels and expert attention.

Prior efforts to integrate eye-gaze into DL have largely focused on a single reader or pooled gaze across readers into a unified map [15-18]. For example, GazeGNN fuses gaze with feature-graph CXR representations to guide model attention and improve classification [15]. These approaches treat gaze as a single "expert signal" (one reader or an averaged map), implicitly assuming experts are interchangeable. This conceals a question for radiologist-informed AI: how the number and combination of radiologists and residents affect downstream performance.

Studies in observer performance and skill assessment have shown that radiologists differ systematically in visual strategies, cognitive workflows, and diagnostic skills [19, 20]. Multi-reader CXR datasets used for severity assessment have highlighted substantial variability in data labeling and confidence across readers and institutions [21, 22]. ET from different radiologists is not interchangeable noise; it encodes complementary perspectives whose value depends on reader selection and integration. To our knowledge, no prior work has systematically varied the number and diversity of radiologist readers to measure the effect on model performance. We therefore also ask whether gaze-derived gains survive a change of cohort, which requires evaluating every strategy on data from other institutions without retraining.

To address this gap, we introduce GazeRIB (Gaze-guided Radiologist-Informed Blackout), a framework that uses ET-derived blackout masks to systematically evaluate multi-reader ET strategies for lung disease severity classification, using COVID-19 as a well-characterized testbed. We collected ET data from eight experts who repeatedly reviewed the same CXRs, generated blackout CXRs by masking non-gazed regions while preserving attended areas, and used these ET-augmented CXRs in a two-stage pipeline with seven image-classification networks (backbones) pre-trained on ImageNet [23] (DenseNet-121 [24], ResNet-18/50 [25], EfficientNet-B0/B1 [26], Vision Transformer (ViT) [4], and Swin Transformer (Swin) [27]). In Stage 1, the full backbone and head are re-trained on blackout CXRs under five reader-panel strategies, defined in Results. In Stage 2, the backbone is frozen and a new head is fine-tuned on the original CXRs. Labels, architectures and hyperparameters are held fixed across strategies, so differences track reader composition and the masked-image volume it implies.

This is a controlled study of gaze composition rather than a new gaze-aware classifier: every ET strategy is evaluated against a matched per-backbone baseline under identical splits, optimizers and schedules, so differences are attributable to reader-gaze composition under a fixed gaze-guided pipeline. It complements prior gaze-aware architectures such as GazeGNN [15] and RadioTransformer [17], which introduce new classifier structures on top of gaze signals.

Across all backbones, 10 seeds and stratified validation/test splits, the gain from adding readers was concentrated between one reader and three. All8-ET achieved the largest mean gains among the gaze strategies, with per-architecture F1 peaks up to 14.8 percentage points. The three multi-reader subsets also separated from baseline on five of seven metrics after Bonferroni correction, gaining 5.3-5.9 percentage points in F1, whereas single-reader training separated on none in macro F1; three readers recovered 73-80% of the eight-reader gain. Inference-only external validation on two independent cohorts (SIIM-FISABIO-RSNA COVID-19 [28], N = 6,054 and MIMIC-Eye [14] subset of MIMIC-CXR [29], N = 1,938) showed that every multi-reader panel improved cross-cohort transfer in the area under the receiver-operating-characteristic curve (ROC-AUC) on SIIM-FISABIO-RSNA (+1.93 to +4.24 pp), whereas the single-reader average did not change (+0.03 pp) and only Res2 improved that cohort individually (+2.02 pp, uncorrected). On MIMIC-Eye no panel differed significantly from baseline (All8-ET +0.75 pp, $p = 0.054$), whereas the single-reader average declined (-1.40 pp, $p = 0.008$). The number of contributing readers therefore changes both what the model learns and how far that learning transfers.

# 2 Results

## 2.1 Study cohorts and reader groups

We evaluated GazeRIB on 2,897 frontal CXRs from 1,366 patients treated at Emory Healthcare (Table 1). The task was four-class severity prediction (0 = normal to 3 = severe), consistent with prior COVID-19 severity work [21, 22, 30]. Patients were 48.3% male (mean age 58.5 ± 17.6 years) and predominantly African

American or Black (73.6%); see Table 1.

**Table 1 Demographic characteristics of the 2,897 frontal CXRs from 1,366 patients in the evaluation cohort, stratified by four-class lung disease severity (0-3). Age and body mass index (BMI) are reported as mean ± standard deviation (SD); sex and race are percentages (top three race categories plus "Other/Unknown"). Demographics were available for 2,699 of 2,897 CXRs (93.2%); BMI was recorded for 1,269 (43.8%), so BMI statistics describe that subset**

| Category | Number of CXRs | Age (mean ± SD) | BMI (mean ± SD) | Male (%) | Female (%) | African American or Black (%) | Caucasian or White (%) | Asian (%) | Others or Unknown (%) |
|---|---|---|---|---|---|---|---|---|---|
| Severity = 0 | 278 | 53.4 ± 20.9 | 29.9 ± 9.7 | 48.7 | 51.3 | 78.7 | 16.0 | 2.3 | 3.0 |
| Severity = 1 | 907 | 59.3 ± 18.4 | 31.0 ± 8.3 | 50.6 | 49.4 | 77.6 | 14.0 | 0.8 | 7.6 |
| Severity = 2 | 703 | 61.1 ± 15.5 | 32.8 ± 8.7 | 53.1 | 46.9 | 71.0 | 14.5 | 3.0 | 11.5 |
| Severity = 3 | 1,009 | 58.7 ± 15.6 | 36.3 ± 10.3 | 52.7 | 47.3 | 67.1 | 17.5 | 2.8 | 12.6 |

Severity labels for all 2,897 CXRs were assigned before ET using the same four-class severity scale, in two labeling efforts: 2,120 CXRs were scored by three readers and the remaining 777 by eight. The image-level reference label is the median score across the readers who scored that CXR; the source study reports 77% of studies agreeing within one severity grade [21]. The resulting class distribution was 9.6% normal (278 CXRs), 31.3% mild (907 CXRs), 24.3% moderate (703 CXRs), and 34.8% severe (1,009 CXRs). None of these labeling readers was among the eight experts (five board-certified radiologists and three residents) whose eye movements were recorded.

ET data were collected in a prior AI-assisted observer-performance study, in which eight experts each diagnosed the same 80 CXRs (one per patient) up to nine times over approximately three months [20, 31]. The present work uses only the raw CXRs and the ET recordings, discarding all AI overlays and user-interface (UI) elements from the prior study; none of the eight ET readers labeled the 2,897-CXR evaluation cohort, and all 80 ET CXRs are disjoint from that cohort. Patient demographics of the 80-CXR ET subcohort and the full quality-control protocol are reported in Methods (ET observer study).

After ET data quality control (see Methods, ET preprocessing and blackout mask generation), each expert contributed 7-9 valid reading sessions per CXR, yielding 680-720 ET-augmented CXRs per reader (details in Methods). We denote the five radiologists as Rad1-Rad5 and the three residents (each with at least one year of training remaining) as Res1-Res3.

## 2.2 Gaze separates from the no-gaze baseline only when pooled

We compared five ET strategies against the per-model baseline (details in Methods):

1. Single-ET: GazeRIB training using one expert's ET data at a time, yielding 680-720 ET-augmented CXRs per single-ET dataset (8 datasets, one per expert).
2. 3Res-ET: combining ET-augmented CXRs from all three residents into the same dataset, yielding 2,120 ET-augmented CXRs.
3. 5Rad-ET: combining ET-augmented CXRs from all five radiologists into the same dataset, yielding 3,565 ET-augmented CXRs.
4. Mix5-ET: combining ET-augmented CXRs from mixed-seniority subset of four radiologists (Rad1-Rad4) and one resident (Res1), yielding 3,541 ET-augmented CXRs.
5. All8-ET: combining ET-augmented CXRs from all eight experts into the same dataset, yielding 5,685 ET-augmented CXRs.

Rather than enumerate every reader subset, we chose five strategies that span clinically meaningful combinations: Single-ET probes individual-reader informativeness; 3Res-ET and 5Rad-ET contrast homogeneous resident-only and radiologist-only supervision; Mix5-ET (4 radiologists + 1 resident) adds a mixed-seniority subset at the same five-reader count as 5Rad-ET to isolate seniority-mix effects from reader-count effects; and All8-ET is the maximal-diversity setting.

Gaze-guided training improved performance relative to the per-model baselines. Across all backbones and 10 seeds, every multi-reader strategy exceeded the baseline on all seven metrics and Single-ET on all but precision, though only the panels separated statistically (Supplementary Tables 1 and 2, Fig. 1).

Changes are mean percentage-point (pp) differences (Δ) with 95% confidence intervals, from two-sided paired Wilcoxon signed-rank tests with Bonferroni-adjusted p values ($p_{adj}$) and Cohen's standardized effect size for paired data ($d_z$) [32-34] (see Methods, Comparative statistics).

Relative to the baseline, All8-ET gained +7.47 percentage points (pp) in F1, +5.90 pp in ROC-AUC and +6.20 pp in AU-PRC (Supplementary Table 1 and Table 2). These gains reflect concurrent improvements in recall (+5.8 pp) and precision (+5.7 pp) without loss of specificity (Supplementary Table 2), indicating that All8-ET reduces both false negatives and false positives.

Recall, precision and specificity followed suit (Supplementary Table 2, Fig. 1): All8-ET improved all three (Table 2).

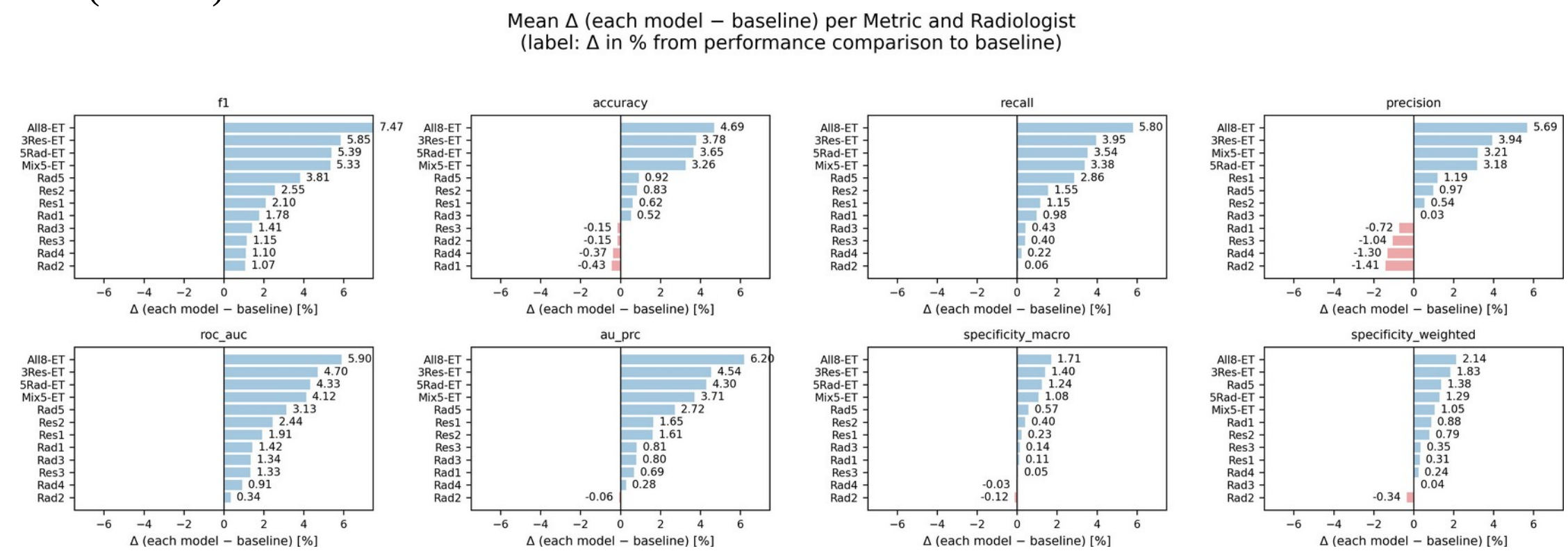

**Fig. 1 Multi-metric performance changes with gaze-guided training. Each panel reports one metric, in the order F1, accuracy, recall, precision, ROC-AUC, AU-PRC, macro specificity and weighted specificity. Bars give the mean change relative to the per-model baseline, averaged over 10 random seeds and 7 backbone architectures, so the baseline is the zero line and not a bar. Within every panel the twelve bars are the four multi-reader strategies, 3Res-ET (three residents), 5Rad-ET (five radiologists), Mix5-ET (four radiologists and one resident) and All8-ET (all eight experts), together with the eight Single-ET models trained on one reader each (Rad1 to Rad5 and Res1 to Res3). Bar values and axes are in percentage points, shown as [%] on the axis label. Accuracy is shown for reference and was not included in the multiplicity correction.**

## 2.3 Multi-reader ET outperforms single-reader strategies

The number and combination of readers' gaze in the GazeRIB framework influenced the performance gain. The decisive factor was whether the gaze came from one reader or from several (Fig. 1 and Table 2). Across seven architectures, Single-ET improved macro F1 by +1.87 pp and ROC-AUC by +1.60 pp, neither of which reached $p_{adj} < 0.05$. Every multi-reader strategy did: 3Res-ET gained +5.85 pp in F1, +4.70 pp in ROC-AUC and +4.54 pp in AU-PRC, 5Rad-ET +5.39, +4.33 and +4.30 pp, and Mix5-ET +5.33, +4.12 and +3.71 pp, all with $p_{adj} \leq 0.0121$ and effect sizes $d_z$ between 0.43 and 0.54. Mix5-ET behaved comparably to the two homogeneous subsets, although one panel per configuration cannot separate reader number from seniority. Macro F1 gained +3.98 pp from one reader to three and a further +1.62 pp from three to all eight. Three readers delivered 73-80% of the eight-reader gain across F1, ROC-AUC and AU-PRC. Measured on the 3,150 unassisted reading masks, one reader retains 20.6% of the image and two readers overlap by a Jaccard index [35] of 0.411; pooling all eight raises the coverage available across training images to 45.1%, less steeply than performance rises. The AI-assisted readings give lower coverage, 10.3% for one reader, and the ordering holds across reading subset and fixation threshold (Supplementary Table 9). These intervals describe model-run variation. A separate cohort bootstrap, resampling by patient rather than image, left the All8-ET macro F1 interval essentially unchanged (95% CI 4.74 to 10.22 pp by patient versus 4.60 to 10.16 pp by image).

Relative to Single-ET, All8-ET added +5.6 percentage points in F1, +4.3 in ROC-AUC, and +5.1 in AU-PRC (Table 2, Fig. 1). Against the no-gaze baseline, All8-ET separated on all seven metrics after Bonferroni correction ($p_{adj} \leq 0.02$) with moderate effect sizes (Cohen's $d_z$ 0.4 to 0.7). Because All8-ET also pools the largest blackout-CXR volume, reader-diversity and training-volume effects are coupled in this design (Discussion). Architecture- and metric-level breakdowns are provided in Supplementary Tables 3 and 4.

**Table 2 Strategy-level gains over the per-model baseline: For each ET strategy, we report mean change (Δ in percentage points across architectures and seeds), 95% confidence interval (CI), paired two-sided Wilcoxon p value, Bonferroni-adjusted $p_{adj}$, and Cohen's $d_z$, over n = 70 paired architecture and seed observations per strategy and metric. Values with $p_{adj}$ < 0.05 after Bonferroni correction over all 35 strategy-metric tests are in bold. All four multi-reader strategies differ from the baseline on F1, ROC-AUC, AU-PRC, recall and macro specificity, whereas single-reader training (Single-ET) does not differ from the baseline on any metric.**

| **Strategy** | **ΔF1 [95% CI]** | **$p$ / $p_{adj}$ / $d_z$** | **ΔROC-AUC [95% CI]** | **$p$ / $p_{adj}$ / $d_z$** |
|---|---|---|---|---|
| Single-ET | 1.87 [−0.21, 3.84] | 0.0838 / 1.0000 / 0.21 | 1.60 [0.20, 2.93] | 0.0118 / 0.4115 / 0.27 |
| 3Res-ET | **5.85 [3.11, 8.45]** | **0.0002 / 0.0064 / 0.49** | **4.70 [2.75, 6.68]** | **<0.0001 / 0.0012 / 0.54** |
| 5Rad-ET | **5.39 [2.27, 8.21]** | **0.0002 / 0.0065 / 0.43** | **4.33 [1.91, 6.53]** | **0.0002 / 0.0082 / 0.44** |
| Mix5-ET | **5.33 [2.67, 7.96]** | **0.0002 / 0.0058 / 0.48** | **4.12 [1.95, 6.20]** | **0.0001 / 0.0052 / 0.44** |
| **All8-ET** | **7.47 [4.60, 10.16]** | **<0.0001 / 0.0002 / 0.62** | **5.90 [3.89, 7.91]** | **<0.0001 / <0.0001 / 0.67** |
| **Strategy** | ΔAU-PRC [95% CI] | $p$ / $p_{adj}$ / $d_z$ | **ΔRecall [95% CI]** | $p$ / $p_{adj}$ / $d_z$ |
| Single-ET | 1.06 [−0.44, 2.45] | 0.0692 / 1.0000 / 0.17 | 0.96 [−0.53, 2.31] | 0.1364 / 1.0000 / 0.16 |
| 3Res-ET | **4.54 [2.61, 6.45]** | **<0.0001 / 0.0010 / 0.52** | **3.95 [2.08, 5.73]** | **0.0001 / 0.0026 / 0.50** |
| 5Rad-ET | **4.30 [2.07, 6.38]** | **0.0001 / 0.0038 / 0.46** | **3.54 [1.41, 5.52]** | **0.0006 / 0.0196 / 0.40** |
| Mix5-ET | **3.71 [1.64, 5.71]** | **0.0003 / 0.0121 / 0.41** | **3.38 [1.43, 5.24]** | **0.0004 / 0.0124 / 0.42** |
| **All8-ET** | **6.20 [4.17, 8.19]** | **<0.0001 / <0.0001 / 0.68** | **5.80 [3.75, 7.80]** | **<0.0001 / <0.0001 / 0.66** |
| **Strategy** | **ΔPrecision [95% CI]** | $p$ / $p_{adj}$ / $d_z$ | **$\Delta Spec_{macro}$ [95% CI]** | $p$ / $p_{adj}$ / $d_z$ |
| Single-ET | −0.22 [−2.20, 1.80] | 0.9184 / 1.0000 / −0.02 | 0.17 [−0.30, 0.60] | 0.4662 / 1.0000 / 0.09 |
| 3Res-ET | 3.94 [1.17, 6.57] | 0.0043 / 0.1488 / 0.33 | **1.40 [0.76, 2.06]** | **0.0002 / 0.0059 / 0.50** |
| 5Rad-ET | 3.18 [−0.21, 6.22] | 0.0148 / 0.5177 / 0.23 | **1.24 [0.54, 1.90]** | **0.0008 / 0.0282 / 0.42** |
| Mix5-ET | 3.21 [0.20, 6.16] | 0.0158 / 0.5522 / 0.25 | **1.08 [0.45, 1.71]** | **0.0014 / 0.0494 / 0.40** |
| **All8-ET** | **5.69 [2.63, 8.54]** | **0.0003 / 0.0092 / 0.44** | **1.71 [1.06, 2.34]** | **<0.0001 / 0.0002 / 0.61** |
| **Strategy** | **$\Delta Spec_{weighted}$ [95% CI]** | $p$ / $p_{adj}$ / $d_z$ | | |
| Single-ET | 0.46 [−0.61, 1.44] | 0.4734 / 1.0000 / 0.10 | | |
| 3Res-ET | 1.83 [0.63, 2.98] | 0.0052 / 0.1820 / 0.34 | | |
| 5Rad-ET | 1.29 [−0.02, 2.54] | 0.0179 / 0.6273 / 0.23 | | |
| Mix5-ET | 1.05 [−0.28, 2.31] | 0.0767 / 1.0000 / 0.18 | | |
| **All8-ET** | **2.14 [0.98, 3.26]** | **0.0005 / 0.0180 / 0.42** | | |

## 2.4 Gaze informativeness varies by reader

Each reader was also tested individually against the baseline (Supplementary Table 8). To understand why multi-reader ET is beneficial, we examined how performance changed when each reader's gaze was used separately. Single-ET models trained on each expert separately improved F1, ROC-AUC and AU-PRC for nearly all readers, but the magnitude of the gains differed (Fig. 2). Rad5 and Res2 showed numerically larger improvements than other readers, with F1 gains of +3.81 and +2.55 pp and similar gains in ROC-AUC and AU-PRC, although individual reader rankings were not tested pairwise. Some readers such as Rad2 showed smaller or slightly negative changes in precision and specificity while still improving recall.

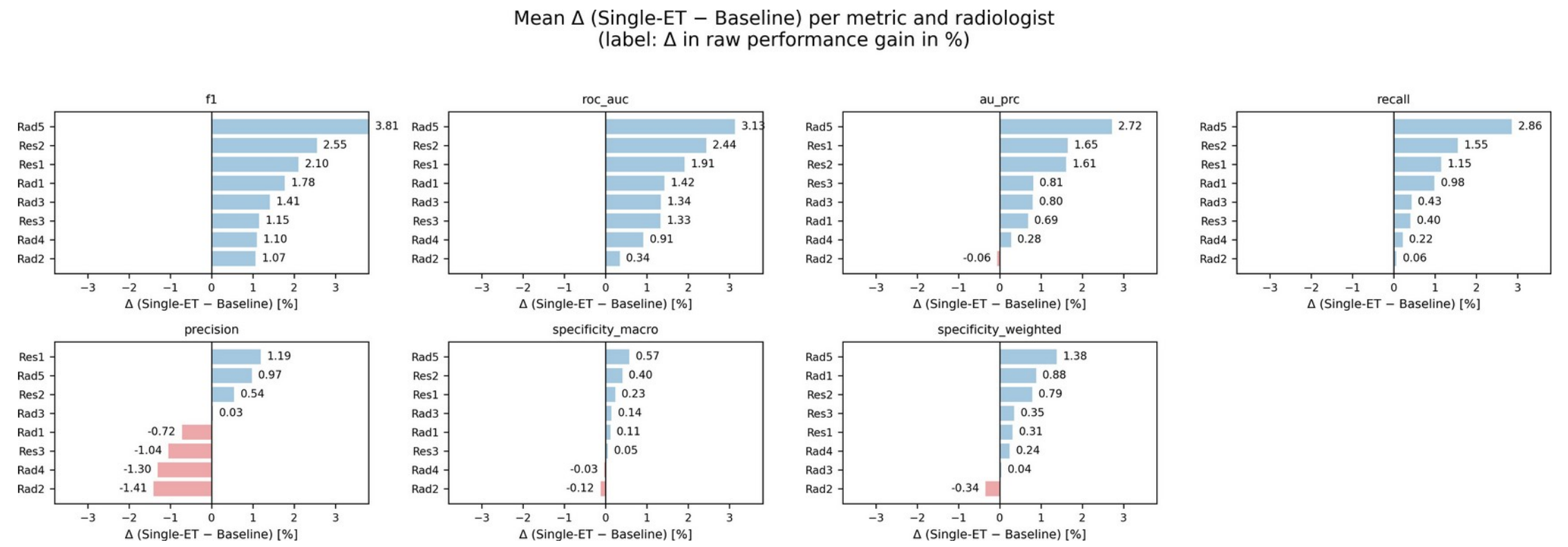

**Fig. 2 Single-reader gaze effects across metrics: Mean change in performance (in percentage points) for Single-ET models trained with gaze from each individual expert, relative to the per-model baseline, across 10 random seeds and 7 backbone architectures. The seven panels are, in order, F1, ROC-AUC, AU-PRC, recall, precision, macro specificity and weighted specificity; accuracy is not shown. Axis values are in percentage points, shown as [%] on the axis label. These are the same eight Single-ET models that appear among the bars of Fig. 1, redrawn on their own scale.**

## 2.5 Convolutional backbones gain most, transformers least

We then assessed how strongly gaze-guided training affected different neural network architectures. Convolutional backbones showed the largest gains from All8-ET (Supplementary Figs. 1-7). Compared with baseline, All8-ET increased F1 by 10.4-14.8 pp for DenseNet-121, EfficientNet-B0, ResNet-18 and ResNet-50, with AU-PRC gains of 8.4-12.8 pp and accompanying improvements in recall and ROC-AUC. EfficientNet-B1 benefited less (+7.8 pp F1).

In contrast, transformer-based backbones showed mixed or negative effects. Swin was essentially unchanged under All8-ET (ΔF1 +0.06 pp) but lost ground under every subset strategy. Similarly, ViT consistently degraded on all seven metrics, among them F1, AU-PRC, and both specificity metrics when trained with ET-based blackout (Supplementary Figs. 1-7).

## 2.6 Pooled gaze also improves binary calibration

Calibration improved with gaze guidance, and improved most when gaze was pooled (Supplementary Fig. 8, Supplementary Table 5). On the clinically actionable normal-versus-abnormal collapse, expected calibration error (ECE) [36] fell from 0.142 for the no-gaze baseline to 0.116 for Single-ET and to 0.069-0.080 for the four multi-reader panels, so pooling reduced miscalibration about 2.5 times more than a single reader did. Brier score [37] followed the same ordering (0.103 baseline, 0.096 Single-ET, 0.085-0.087 multi-reader). A constant predictor emitting the cohort base rate attains ECE 0.000 by construction, and Brier 0.087, with no discriminative value, so these values must be read alongside discrimination. On the four-class task the gaze gains were smaller (ECE_multi 0.301 baseline versus 0.275-0.294; Brier_multi 0.182 versus 0.170-0.173), so the probability benefit concentrates on the binary distinction that drives triage.

## 2.7 Panel gaze improves transfer on SIIM-FISABIO-RSNA and holds parity on MIMIC-Eye

To assess generalizability beyond Emory, we performed inference-only external validation on two public datasets. Both use labeling schemas different from the internal four-class task, so they probe cross-task transfer rather than replicating the internal endpoint. The first was the SIIM-FISABIO-RSNA 2021 COVID-19 cohort [28] (N = 6,054 studies, 6,334 images; 55.7% male; binary label: 0 = Negative for Pneumonia, 1 = any of Typical/Indeterminate/Atypical Appearance; 72.3% abnormal). The second was the MIMIC-Eye [14] subset of MIMIC-CXR [29] (N = 1,938 images; binary labels from CheXpert [3]: No Finding vs. lung opacity/consolidation/pneumonia). All 910 Emory-trained checkpoints (13 configurations × 7 architectures × 10 seeds) were evaluated without weight updates; four-class outputs were collapsed as P(abnormal) = P(severity ≥ 1). Details are in Methods (External validation datasets and inference protocol).

On SIIM-FISABIO-RSNA, multi-reader training produced a clear cross-cohort gain. All8-ET achieved ROC-AUC = 73.0 ± 11.8%, outperforming baseline (68.8 ± 9.2%; Δ = +4.24 pp, paired Wilcoxon $p < 0.001$ uncorrected; Cohen's $d_z = 0.40$), with concordant gains in AU-PRC (+2.13 pp) and accuracy (+7.88 pp). The F1 gain (+9.22 pp) falls to +5.41 pp excluding ViT, whose thresholded metrics are degenerate here (Supplementary Table 6). The three intermediate panels also improved on baseline, by +1.93 to +2.39 pp ($p \leq 0.004$ uncorrected). The single-reader average did not (Single-ET ROC-AUC = 68.8 ± 9.3%; Δ = +0.03 pp, $p = 0.854$), and only Res2 improved individually (+2.02 pp, $p = 0.014$).

On MIMIC-Eye, architecture-specific results mirrored internal Emory findings (Supplementary Table 7): ResNet-18, EfficientNet-B0, DenseNet-121 and ResNet-50 gained ROC-AUC over baseline (+5.42, +4.00, +2.91 and +0.72 pp), while EfficientNet-B1, ViT and Swin declined. The benefit for smaller CNNs therefore persists beyond the training distribution. On MIMIC-Eye All8-ET alone exceeded baseline, whereas on SIIM-FISABIO-RSNA every panel did (Table 3).

**Table 3 External validation: binary classification performance (normal vs. abnormal) on two independent datasets via inference only. ROC-AUC (%) is reported as mean ± SD across 7 architectures and 10 random seeds (N = 70 paired models per strategy for the main strategy rows; SIIM-FISABIO-RSNA N = 6,054 studies / 6,334 images, MIMIC-Eye N = 1,938 images). Δ denotes mean percentage-point (pp) difference from the per-model baseline. p values are from two-sided paired Wilcoxon signed-rank tests and are not corrected for multiplicity; * marks uncorrected p < 0.05. Both cohorts were computed from Stage-2 checkpoints trained at learning rate 5 x 10-4, the same rate as the per-model baseline and the in-domain analysis, so every delta reported here isolates the training strategy rather than a learning-rate difference. On SIIM-FISABIO-RSNA all four panel differences reach p ≤ 0.004, no single-reader difference reaches that threshold, and one reader reaches p < 0.05 (Res2, p = 0.014); on MIMIC-Eye the only differences reaching p < 0.05 are declines, namely the Single-ET average and two individual readers.**

| | SIIM-FISABIO-RSNA (N = 6,054) | | | MIMIC-Eye (N = 1,938) | | |
|---|---|---|---|---|---|---|
| **Strategy** | **ROC-AUC (%)** | **Δ (pp)** | **p** | **ROC-AUC (%)** | **Δ (pp)** | **p** |
| Baseline (no ET) | 68.8 ± 9.2 | - | - | 67.4 ± 7.0 | - | - |
| Single-ET (avg.) | 68.8 ± 9.3 | +0.03 | 0.854 | 66.0 ± 7.5 | -1.40 | 0.008* |
| 3Res-ET | 70.7 ± 11.4 | +1.93 | 0.003* | 67.0 ± 8.9 | -0.45 | 0.759 |
| 5Rad-ET | 70.8 ± 11.4 | +1.96 | 0.004* | 66.5 ± 10.2 | -0.95 | 0.642 |
| Mix5-ET | 71.2 ± 10.9 | +2.39 | 0.003* | 66.9 ± 9.4 | -0.58 | 0.984 |
| All8-ET | 73.0 ± 11.8 | +4.24 | <0.001* | 68.2 ± 9.5 | +0.75 | 0.054 |

# 3 Discussion

This study asks how the composition of the reader panel whose gaze is collected changes the resulting classifier, holding architectures, data, loss, optimizer and schedule fixed across every strategy. The step from one reader to three is larger than the step from three to eight (Table 2). Every multi-reader panel separated from the no-gaze baseline (in macro F1, 5.33-7.47 pp, all $p_{adj} < 0.007$) while single-reader gaze separated on none, and the five-reader panels sat no higher than the three-reader panel. In-domain the decisive factor is whether gaze is pooled at all; under a change of cohort, panel size also matters. With one panel per size this is a pattern across the tested configurations rather than a fitted dose-response curve.

Masking discards diagnostic signal, so the usable question is not whether gaze helps but how many readers it takes to stop discarding it. Reader-level analysis indicates why pooling helps (Fig. 1, Table 2). Most experts improved F1 and AU-PRC on their own, but the magnitude and direction varied by metric and reader (Fig. 2), consistent with long-standing evidence of heterogeneity in radiologists' search strategies [9, 11]. All8-ET improved every metric and outperformed the best single-reader models, suggesting that no single expert's gaze captures the full spectrum of diagnostically relevant regions. Volume alone does not account for this ordering: 5Rad-ET carries 68% more blackout images than 3Res-ET yet gained less (+5.39 versus +5.85 pp), and swapping one reader between Mix5-ET and 5Rad-ET at matched volume changed F1 by 0.06 pp (+5.33 versus +5.39 pp).

The benefits of gaze-guided training are architecture-dependent. Convolutional backbones (DenseNet-121, ResNet-18/50, EfficientNet-B0/B1) gained the most from All8-ET, with F1 improvements of 7.8-14.8 percentage points and consistent gains in AU-PRC, recall and ROC-AUC (Supplementary Table 3, Supplementary Figs. 1-7). These models appear to use gaze-guided blackout as a spatial constraint that concentrates learned features on regions experts consistently scrutinized. Direct causal verification would require sensitivity analyses over mask design. Transformer backbones behaved differently: Swin showed small, noisy changes and ViT degraded across F1, AU-PRC and specificity. Our ET-augmented set contains thousands of blackout CXRs derived from only 80 source images, sufficient for CNNs but below the scale at which high-capacity transformers typically benefit. Within this regime, GazeRIB with All8-ET still yields consistent gains over strong ImageNet-pretrained baselines across all convolutional architectures.

External validation further characterizes scope. All8-ET improved ROC-AUC on SIIM-FISABIO-RSNA (+4.24 pp, $p < 0.001$) and exceeded baseline on MIMIC-Eye (68.2% vs. 67.4%, +0.75 pp, $p = 0.054$). The three smaller panels also improved ROC-AUC on SIIM-FISABIO-RSNA (+1.93 to +2.39 pp, $p \leq 0.004$) and changed by -0.45 to -0.95 pp on MIMIC-Eye, none significantly. The single-reader average improved neither cohort: all eight readers fell below baseline on MIMIC-Eye (average 66.0%), and on SIIM-FISABIO-RSNA the single-reader average in ROC-AUC was indistinguishable from baseline (+0.03 pp, $p = 0.854$). Individually, only Res2 improved SIIM-FISABIO-RSNA ROC-AUC (+2.02 pp, uncorrected; Supplementary Table 10).

The two cohorts agree at the extremes and diverge in the middle. All8-ET is the strongest arm and single-reader gaze the weakest on both, but the intermediate ROC-AUC ordering reverses, Mix5-ET above 5Rad-ET above 3Res-ET on SIIM-FISABIO-RSNA against 3Res-ET above Mix5-ET above 5Rad-ET on MIMIC-Eye, and the signs differ, with four of four panels above baseline on SIIM-FISABIO-RSNA and one of four on MIMIC-Eye. Panel size therefore orders transfer only at the extremes.

Reader identity is a training variable whose effect did not transfer. The reader with the largest in-domain gain, Rad5 at +3.81 pp in macro F1, ranked last of eight on both external cohorts, at -1.55 and -2.37 pp in ROC-AUC (Supplementary Table 10). In-domain benefit did not predict transfer (Spearman rho -0.310 and -0.024, neither significant). This exploratory comparison does not support selecting one reader in place of a panel.

We used COVID-19 severity as a testbed for its well-characterized four-class scale, large multi-reader-labeled cohort, and an independent ET study on the same scale [20-22]. It is therefore a proxy for general lung opacification rather than a COVID-19-specific endpoint. The best COVID-specific systems use larger multi-center cohorts and multimodal inputs [21, 22, 30]. We do not claim state-of-the-art disease classification.

Several constraints merit note. First, four-class severity labels carry intrinsic inter-reader disagreement (60-80% pairwise agreement [20-22]), which caps the achievable ceiling. Second, the ET dataset comprises 80 source CXRs read repeatedly by eight experts, yielding 5,685 reader-session blackout images; standard in observer-performance studies [20]; larger ET cohorts remain an extension. Third, the 80 ET CXRs span all four severity classes, quality strata, and cases a prior COVID-19 model got right and wrong [20]; the external cohorts are not so curated. Fourth, the blackout transform weights fixations by duration, pupil diameter and hit proportion with two-stage Gaussian smoothing, in line with prior ET-guided methods [15-18]. Fifth, ET came from unassisted and AI-assisted reading [20]; overlays were excluded from training images, but residual influence on gaze cannot be excluded. Sixth, the baseline trains only the head whereas every gaze arm updates the full backbone in Stage 1, so gains bundle gaze with added representation learning and cannot separate the two. Head-only adaptation is canonical in low-data transfer, where updating a full backbone from 80 images is prone to overfitting [6, 38]. Seventh, adding readers increases gaze diversity and blackout-image volume together (5,685 versus 711 per reader), a bundling intrinsic to multi-reader pipelines [12-18]. The non-monotonicity above bounds what volume can explain, but sample-count-matched controls would be needed to separate the two. We did not perform subgroup analyses across sex, age or race, so demographic generalization remains open. Finally, we address predictive performance rather than mechanism; attribution-gaze alignment requires targeted follow-up.

The practical implication for gaze collection is that the number of contributing readers is what to record, and that three readers already secure 73-80% of the eight-reader benefit in domain. Every panel we tested also carried a gain to an independent cohort under inference alone, +1.93 to +4.24 pp of ROC-AUC on SIIM-FISABIO-RSNA. Breadth counts for more under a change of institution: on MIMIC-Eye the eight-reader panel was the only arm to stay above baseline. Expert attention is therefore a design variable rather than a fixed resource sampled once: which readers are recorded changes what the model learns, in distribution and beyond it. Larger multi-center ET cohorts would test portability. Saccade features, task-phase information and joint gaze-image modeling may matter most for transformers and for video modalities. The framework extends to other 2D tasks and, in principle, to volumetric imaging.

# 4 Online Methods

## 4.1 Study design, cohorts, and ethical approval

This was a retrospective single-health-system study conducted at Emory Healthcare affiliated hospitals (Atlanta, GA, USA). We used 2,897 frontal (anterior-posterior and posterior-anterior) CXRs from 1,366 patients imaged for COVID-19 between January and November 2020. The study was approved by the Institutional Review Board (IRB), with a waiver of informed consent due to the retrospective use of de-identified data.

The 2,897-CXR cohort was used exclusively for validation and testing of all models (baseline and gaze-guided). Eye-tracking data were collected in an independent observer study on a different dataset of 80 CXRs using the same four-class severity scoring system but with a different set of eight experts [20]. None of the eight ET readers participated in labeling the 2,897-CXR cohort. Disjointness between the two cohorts was verified during data cleaning by patient-identifier cross-check: no patient in the 80-CXR ET cohort appeared in the 2,897-CXR evaluation cohort, so validation and test evaluations are strictly held out with respect to the ET training data.

## 4.2 Severity labeling and clinical task

The clinical task was four-class lung disease severity prediction (0 = normal, 1 = mild, 2 = moderate, 3 = severe), consistent with prior work on COVID-19 severity scoring in this line of research [21, 22]. For the 2,897-CXR evaluation cohort used in this study, severity scores were assigned in two labeling efforts: 2,120 CXRs were scored by three readers and the remaining 777 by eight. The image-level reference label is the median score across the readers who scored that CXR [21]. These labels served as the reference standard for all validation and test evaluations reported in this work.

Inter-reader variability on this four-class scale has been characterized in a previous study [21]. In that analysis, most CXRs received identical or near identical scores across readers, with 77% of the studies showing at most a one-grade difference between readers. Therefore, the median label represented a high-agreement consensus while still reflecting residual uncertainty in borderline cases.

For the ET observer study [20], the same four-class scale was used but with a separate panel of eight readers (five radiologists and three residents). The ET study characterized observer performance and gaze behavior, and also established reference severity labels for these 80 CXRs.

## 4.3 ET observer study

For the ET cohort used in this work, we used ET data collected in our prior IRB-approved AI-assisted observer-performance study on COVID-19 chest radiography [20]. In the study [20], 80 representative frontal CXRs (20 per severity class) were selected to span a range of COVID-19 pathological appearances, image quality and diverse patient demographics, and to include cases where a previously developed COVID-19 severity model [21] was both correct and incorrect in its predictions. As described in the Results section, these cases were read by eight experts, consisting of five board-certified radiologists and three radiology residents who had at least one year of training remaining.

Each expert viewed the same set of 80 CXRs across multiple sessions, experiencing five total viewing conditions: one unassisted condition (CXR only) and four AI-assisted conditions (CXR with a GradCAM heatmap, CXR with a categorical severity label, CXR with a bar plot of severity probabilities, and CXR with both a heatmap and a bar plot) in which an AI-assisted output was displayed adjacent to the CXR [20, 21]. The AI-assisted outputs were generated by a previously developed DL model trained for COVID-19 severity [21], which is distinct from the models used in the present study. For each expert and CXR, the unassisted condition was read in five independent sessions (separated by ≥3 weeks to reduce memory effects [20]) and each of the four AI-assisted conditions was read in one session after the unassisted condition, yielding nine total reading sessions from the same setting for each expert and CXR.

In the present work, we only used the CXRs and their corresponding ET data, while ensuring all AI-assisted overlays and outputs from [21] were excluded entirely. We cropped each screen recording to retain only the CXR region before applying the blackout technique. This ensured that our models never saw the original AI-assisted output regions from the previously developed model [21]. Importantly, no ET data were involved in the development of the prior model, which was trained and evaluated on COVID-19 Pneumonia CXRs [21].

The 8 experts viewed the 80 CXRs on a standardized display (Eizo Flexscan EV24510 with resolution 1920 x 1080 and 23.8 inch diagonal), using the Tobii Pro Spectrum ET system. This system used video-based pupil and corneal reflection ET with both dark and bright pupil illumination modes and sampled at 1200 Hz. Two cameras captured stereo images of both eyes for accurate measurement of eye-gaze and eye position in the 3D space, as well as pupil diameter. ET data included a continuous record of gaze position within the space defined by the monitor. This continuous data was post-processed using Tobii software to characterize fixations, dwell times on the images as well as specific targets, total viewing time, location returns, and saccades. Each expert viewed the same set of 80 frontal CXRs (20 per severity class) across several conditions in multiple sessions with at least 3 weeks between sessions to reduce memory effects [20].

Since each of the 8 experts viewed all 80 CXRs nine times (5 unassisted readings + 4 AI-assisted conditions), there should be 720 readings per expert from 80 cases × 9 readings. In practice, not all readings produced usable ET data. We excluded trials with calibration and ET recording failures at the start of a session, or incomplete or prematurely terminated sessions. Because calibration failures and incomplete sessions varied across experts and sessions, the number of usable readings per reader was not the same. After exclusions, readers contributed between 7 and 9 reading sessions per case, resulting in 680-720 valid ET-augmented readings per reader. The exclusions above, not the repetition itself, are what controlled ET data quality.

Raw ET records were exported as tab-delimited files with one row per event. The analysis uses Participant (reader ID), Event type (Fixation or Saccade), FixationPointX/Y (pixels), Duration (ms), Average pupil diameter (mm), Hit proportion and the CXR image name. Hit proportion is the fraction of gaze samples falling inside the display area of interest (AOI), which comes directly from the eye-tracker's event parser. The export also carries timing, validity, saccade-kinematics and area-of-interest fields that this analysis does not use.

## 4.4 ET preprocessing and blackout mask generation

To convert ET into a training signal, we transformed each reader’s fixation pattern on each CXR into a smooth “blackout” mask that preserves attended regions while suppressing non-viewed areas.

For fixation i, the ET logs provided normalized image coordinates in pixels (xi, yi), duration τi (ms), hit proportion (the eye-tracker-derived fraction of gaze samples falling within the image-display region of interest on screen, as defined by the Tobii event parser; this is a recording-quality metric, not an anatomical annotation), and average pupil diameter (mm). We first clipped duration and pupil diameter to physiologically plausible ranges for CXR reading. The per-pixel blackout mask is constructed by the sequence of transformations in equations (1)-(12) below.

$$\tau_i \leftarrow \text{clip}(\tau_i, 150, 300)\,\text{ms}, \overline{d_i} \leftarrow \text{clip}(\overline{d_i}, 2.5, 5.0)\,\text{mm}, \tag{1}$$

and defined a per-fixation intensity weight with hit proportion as

$$\alpha_i = 0.7 + 0.3\,h_i, \tag{2}$$

so that fixations that were longer, more dilated and more reliably recorded would yield stronger preservation of underlying pixels and contribute strongly to the mask.

An adaptive radius was then defined by

$$r_i = 50 + \frac{\tau_i \overline{d_i}}{10}\,(\text{pixels}) \tag{3}$$

ensuring longer, larger-pupil fixations influence wider regions. This sets the spatial extent of each fixation's influence. The given clipped ranges produced ri between 87 and 200 pixels in the same mask coordinate grid as the base radius, which is slightly larger than typical fixation scales at the display resolution used.

For each pixel (*x, y*), we computed the Euclidean distance $d(x_i, y_i)$ to fixation $i$ and defined

$$m_i(x, y) = \alpha_i\, max\left(0, 1 - \frac{d_i(x, y)}{r_i}\right), \tag{4}$$

which decays linearly to zero at distance $r_i$. Summing across all fixations from each individual reading session separately produced an initial attention map

$$M'(x, y) = \sum_i m_i(x, y), \tag{5}$$

which we clipped to the unit interval to avoid oversaturation and overlapping intensities,

$$M''(x, y) = min\left(1, max\left(0, M'(x, y)\right)\right). \tag{6}$$

To blend adjacent regions and mimic parafoveal pooling and smooth transitions between nearby fixations, we applied two Gaussian blurs in sequence. Let Gσ denote convolution with an isotropic 2D Gaussian kernel centered at origin and normalized to sum to 1.

$$G_\sigma(M)(x, y) = (g_\sigma * M)(x, y), g_\sigma(u, v) = \frac{1}{2\pi\sigma^2}\exp\left(\frac{-(u^2 + v^2)}{2\sigma^2}\right). \tag{7}$$

First, a fixed-width blur with standard deviation $\sigma_1$ = 10 pixels yielded

$$M_1 = G_{\sigma_1}(M''). \tag{8}$$

Next, we defined clip(*x, a, b*) = min(max(*x, a*), *b*). Thus, the second pupil-scaled blur is

$$\sigma_2 = \text{clip}(3\overline{d}^{¿}, 10, 30) \tag{9}$$

where $\bar{d}^{*}$ is the session-level mean pupil diameter, bounded to the same range as $\bar{d}_i$. This produced the

final smoothed mask

$$M(x,y)=G_{\sigma_2}(M_1). \tag{10}$$

Finally, we normalized *M* to [0, 1] via

$$\widetilde{M}(x,y)=\frac{M(x,y)-\min_{x',y'} M(x',y')}{\max_{x',y'} M(x',y')-\min_{x',y'} M(x',y')}, \tag{11}$$

where x′ and y′ are dummy variables that run over all pixel locations in the image. We then applied it multiplicatively to the original CXR Iraw(x, y), yielding the ET-augmented CXR

$$I_{\mathrm{ET}}(x,y)=I_{\mathrm{raw}}(x,y)\widetilde{M}(x,y). \tag{12}$$

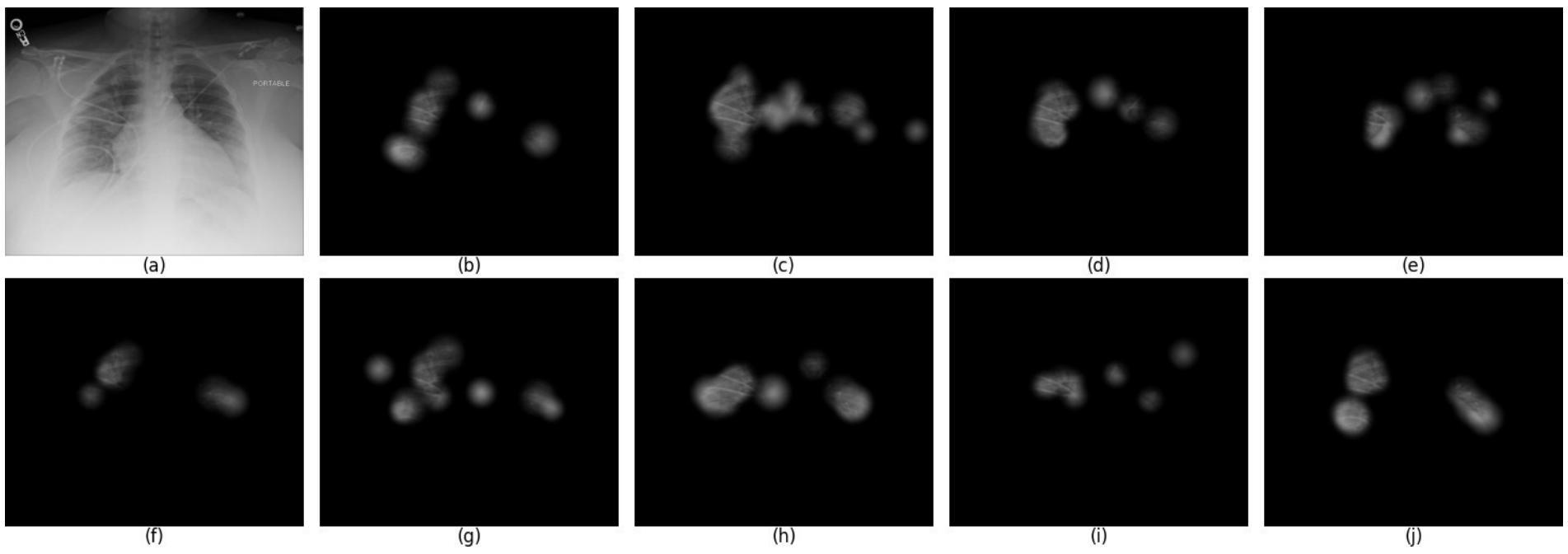

**Fig. 3 Repeated blackout augmentations from a single expert. (a) Original CXR of severity class 2. (b-j) Nine independent blackout overlays derived from separate ET sessions by the same radiologist (Rad1), illustrating within-reader variability in gaze-guided masks.**

Overall, the fixation durations of 150-300 ms and pupil diameters of 2.5-5.0 mm are within expected ranges for medical image interpretation and were used to bound ET events. The base radius (50 pixels in the mask coordinate grid) sets a sensitive transition on the order of a few degrees of visual angle on the ET setup on 23.8 inch, 1920×1080 display, while the two-stage Gaussian smoothing (σ1=10 pixels; σ2 scaled to session-level pupil size) mimics broader parafoveal pooling. Weighting by hit proportion emphasizes fixations that were most consistently recorded within the eye-tracker display AOI. An example of repeated blackout masks from a single expert for the same CXR is shown in Fig. 3.

This procedure yields one blackout CXR per reader and reading session for each of the 80 ET cases, which are then combined into each expert's ET dataset to form the Single-ET, 3Res-ET, 5Rad-ET, Mix5-ET, and All8-ET training sets used in GazeRIB as described in Results (Gaze separates from the no-gaze baseline only when pooled). These datasets differ only in whose gaze is included and how it is combined; the underlying 80 source CXRs and their labels are identical across strategies.

## 4.5 Backbone architectures and classifier heads

We evaluated seven ImageNet-pretrained architectures: DenseNet-121, ResNet-18, ResNet-50, EfficientNet-B0, EfficientNet-B1, a vision transformer (ViT), and a Swin transformer (Swin). For all models, the final task was four-class lung disease severity prediction.

To avoid confounding effects from large, architecture-specific classifier heads, we replaced the original ImageNet classification layers with a lightweight, harmonized head. For the convolutional backbones and ViT, the head consisted of a small multilayer perceptron (MLP) with an intermediate bottleneck (Eq. 13):

$$\text{Dropout}\left(p=0.3\right) \rightarrow \text{Linear}\left(d_{\text{feat}} \rightarrow 256\,\text{or}\,512\right) \rightarrow \text{ReLU} \rightarrow \text{Linear}\left(\cdot \rightarrow 4\right), \tag{13}$$

where $d_{\text{feat}}$ is the backbone feature dimensionality.

## 4.6 Baseline training

The baseline condition quantified performance without any ET information while retaining the lightweight classification head on top of the pretrained backbone. Each backbone (DenseNet-121 [24], ResNet-18 and ResNet-50 [25], EfficientNet-B0 and EfficientNet-B1 [26], ViT [4], and Swin [27]) was initialized from ImageNet-1K [23] pretrained weights and kept frozen. We froze all convolutional/transformer layers and trained only the classification head on the 80 original CXRs (no ET) using the same four-class labels as in the ET study. The backbone is the stack of convolutional or transformer layers that turns an image into a feature vector; the head is the small classifier that maps that vector to the four severity classes. ImageNet pretraining supplies the backbone with general visual features, so only the head has to be learned from the limited labeled CXRs available here. Updating every backbone weight from 80 images instead risks overfitting, which is why adaptation is restricted to the head in this regime [6, 38].

The loss function was focal loss ($\gamma$ = 2) with class weights inversely proportional to training set class frequencies. We used the AdamW optimizer with learning rate $5 \times 10^{-4}$ for the head and weight decay $5 \times 10^{-4}$. A Reduce-on-Plateau scheduler decreased the learning rate by a factor of 0.05 after three epochs without improvement in the validation early stopping metric (monitored over a six-epoch window).

## 4.7 GazeRIB two-stage training

GazeRIB uses a two-stage training procedure to isolate the contribution of gaze-guided representation learning in the backbone versus the pretrained representation and weights as shown in Fig. 4. Stage 2 aligns the classifier with unmasked inputs: it fine-tunes the head on non-blackout CXRs so that the model sees full anatomical context at inference time, matching the distribution of images it will encounter in clinical use and preventing a train-test distribution shift caused by blackout masking during Stage 1.

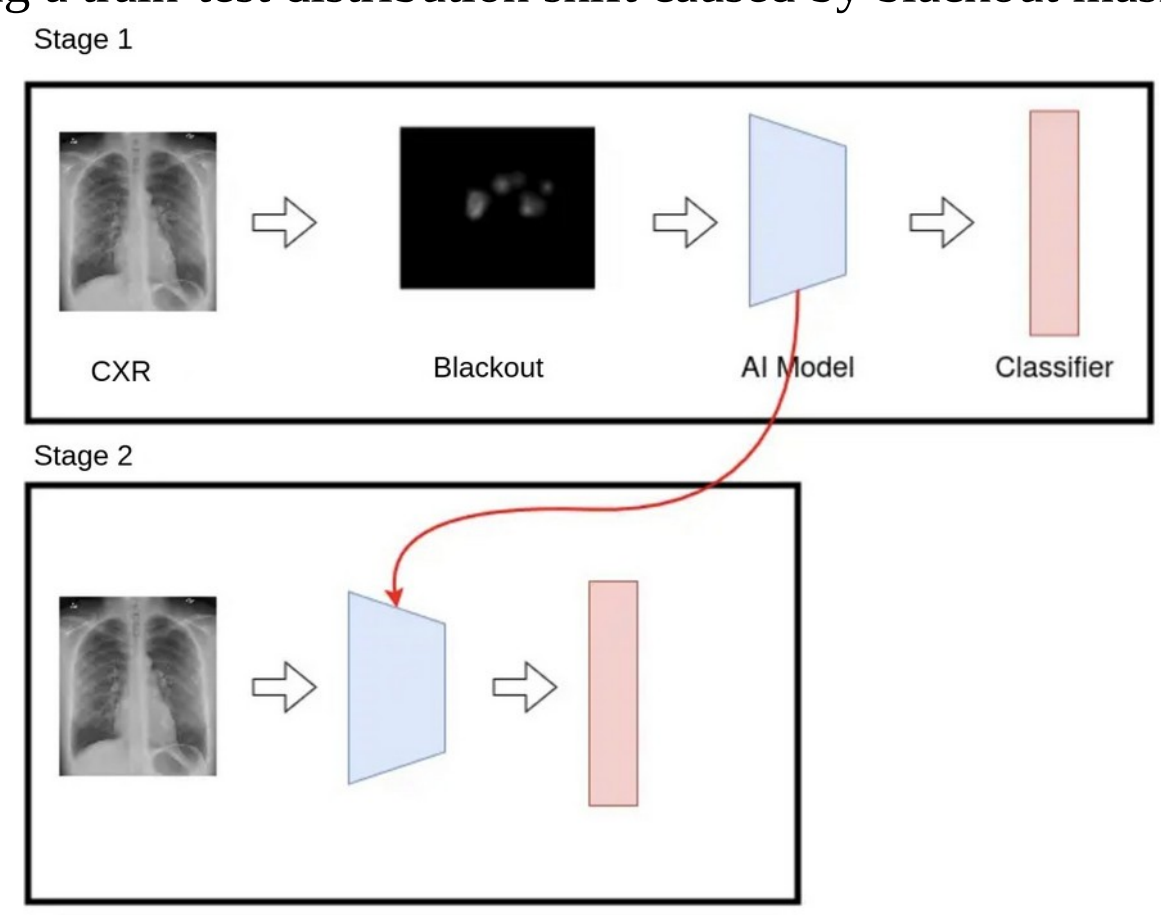

**Fig. 4 Gaze-guided two-stage training in GazeRIB: In Stage 1, radiologist ET is converted into blackout masks that suppress non-viewed regions while preserving attended areas. Blackout CXRs are used to re-train the full backbone and classifier head from ImageNet initialization under different ET strategies. In Stage 2, the backbone is frozen and a new classifier head is fine-tuned on the original CXRs, restoring access to the full image. Within Stage 1 the elements are, from left, the input CXR, the gaze-derived blackout mask, the backbone (AI Model) and the classifier head (Classifier); the red arrow carries the trained backbone into Stage 2, where it is frozen.**

In Stage 1 (gaze-guided re-training of backbone and head), for each backbone and each ET strategy (Single-ET, 3Res-ET, 5Rad-ET, Mix5-ET, All8-ET), we trained the full network on blackout CXRs derived from the original 80 ET source CXRs. Validation and testing used the 2,897-CXR Emory cohort, which excludes the 80 ET source images. Those partitions were drawn at the patient level using the matching-filename identifier, which takes one value per patient (1,366 values for 2,897 CXRs), and stratified by severity class, so every radiograph of a patient fell in the same partition. The split is a deterministic function of the random seed, so Stage 1 and Stage 2 share identical partitions for a given seed and no checkpoint was selected using data reserved for final testing. No blackout image was used for validation or testing at either stage. Training used the same focal loss, class weighting, AdamW optimizer, and Reduce-on-Plateau scheduler as the baseline, with weight decay $5 \times 10^{-4}$. Because Stage 1 updates the backbone, the optimizer used two parameter groups, the backbone at $1 \times 10^{-4}$ and the head at $5 \times 10^{-4}$. Batch size (32), augmentations, and the early-stopping protocol (validation macro-F1 monitored over a six-epoch window) were identical to the baseline condition. The best Stage 1 checkpoint was selected by validation performance under the same random seed.

In Stage 2 (classifier-head fine-tuning on original CXRs), we froze the backbone weights from the best Stage 1 checkpoint and reinitialized the classification head. We then fine-tuned this head on the same 80 original CXRs, using the same labels, data splits, loss, optimizer, and scheduler as in the baseline. This ensured that the only systematic difference between baseline and gaze-guided models was whether the backbone had been exposed to gaze-guided inputs in Stage 1. All models (baseline and ET strategies) were finally evaluated on the 2,897-CXR Emory evaluation cohort using the same stratified validation and test splits for a given seed.

## 4.8 Validation/test splitting and metrics

For each of 10 random seeds, we split the 2,897-CXR evaluation cohort into 20% validation and 80% test sets, stratified by patient and severity class to preserve disease distribution and prevent patient leakage. For each seed, the same validation/test partitions were reused across all architectures and training strategies. Because stratification is applied per seed, test-set sizes differ between seeds, from 2,210 to 2,379 images, so pooled analyses total 162,078 predictions per single-model strategy rather than a round multiple of one split size. During training on the ET-augmented CXRs, a weighted random sampler (based on class frequencies) was used to construct mini-batches among the datasets.

Macro metrics average the four severity grades equally, so the smallest grade counts as much as the largest. For each split we computed macro F1, macro precision and recall, area under the receiver-operating-characteristic curve (ROC-AUC; one-vs-rest), area under the precision-recall curve (AU-PRC), macro specificity (Spec_macro), prevalence-weighted specificity (Spec_weighted), and overall accuracy. We summarize performance as mean ± standard deviation across seeds, and, where indicated, average further across architectures to obtain strategy-level summaries as in Tables 2-3 and Supplementary Tables 1-4.

## 4.9 Comparative statistics

We quantified how each ET strategy changed performance relative to its per-model baseline and examined whether these changes were statistically reliable. For each backbone architecture and random seed, we computed paired performance differences between every gaze-guided strategy and its per-model baseline for each evaluation metric. Let metricstrategy and metricbaseline denote the F1, ROC-AUC, AU-PRC, recall, precision, macro specificity (Specmacro) or prevalence-weighted specificity (Specweighted) obtained by a given ET strategy and by its corresponding baseline model under the same backbone, seed, data splits and training protocol. We defined the per-run performance change (Eq. 14) as

$$\Delta\,\text{metric} = 100 \times \left(\text{metric}_{\text{strategy}} - \text{metric}_{\text{baseline}}\right), \quad (14)$$

so that each backbone and seed pair contributed one paired delta value per strategy and metric. For Single-ET the eight per-reader runs of a given backbone and seed were averaged before pairing, so that strategy also contributes one delta per pair rather than eight. At the strategy level, this yields up to 7 architectures × 10 seeds = 70 paired deltas for each metric. For each strategy and evaluation metric combination, we summarized the distribution of deltas by its sample mean and a 95% CI estimated via non-parametric bootstrap (2,000 resamples over backbone and seed pairs). Observations within the 70-point distribution are not fully independent: different random seeds share the same 80-CXR training cohort and identical data splits within a seed, and different backbone architectures are not drawn from an i.i.d. population of models. The pooled Wilcoxon tests below therefore address whether the paired differences are consistently signed across architecture and seed pairs, and their p values carry that dependence rather than describing a population of models; per-architecture breakdowns (Supplementary Tables 3-4) are provided as a more conservative view of the evidence. Because the evaluation cohort contains repeat images from the same patients, we also recomputed each strategy's macro F1 delta with a cluster bootstrap that resamples held-out patients instead of images (1,093 test patients per seed, 400 resamples per architecture and seed pair). Each run's held-out patient list was read from its own training log, and the record ordering was verified by reconstructing the 2,300-label test sequence and matching it element-wise to the saved predictions. Patient-clustered intervals were close to the image-level ones, indicating that repeat images do not materially inflate the precision of these estimates.

To test whether the performance change differed from zero, we applied a two-sided paired non-parametric test (Wilcoxon signed-rank) to the paired deltas for each strategy and metric, treating the deltas as paired observations relative to each model baseline [32]. Wilcoxon was chosen because it does not assume normality and is robust to potential outliers in the paired-delta distribution, avoiding the need for a prior normality test before inference. The resulting paired p-values quantify evidence against the null hypothesis of symmetric distribution of deltas around zero (median = 0) (no systematic change relative to baseline). Since we tested 5 ET strategies across 7 metrics (35 strategy-metric tests in total), we controlled the family-wise error rate at $\alpha = 0.05$ using Bonferroni correction [33], and Bonferroni adjusted p value as $p_{adj} = \min(p \times 35, 1)$. Therefore, we reported both the paired p-values (Wilcoxon) and the adjusted p-values (Bonferroni), with the pre-specified decision threshold $p_{adj} < 0.05$.

However, to characterize the magnitude of effect, we reported Cohen's $d_z$ for each ET strategy-metric combination, defined as $d_z = \frac{\overline{\Delta}}{s_\Delta}$,where $\Delta$ and $s_\Delta$are the sample mean and standard deviation of the paired deltas, respectively [34]. Thus, $d_z$ expresses the mean performance change in units of across-run variability (with values around 0.2, 0.5 and 0.8 often informally described as small, medium and large effects in the broader effect size literature) [34, 39].

We reported both direction and magnitude of the mean delta and $d_z$, with the paired p values and $p_{adj}$ used to quantify the statistical uncertainty and control for multiple comparisons. Note that all comparative analyses were implemented in Python using SciPy and statsmodels packages.

## 4.10 Calibration analysis

We evaluated how well predicted probabilities reflected the observed outcomes for both the four-class severity scoring task and clinically relevant binary case (any radiographic abnormality, severity ≥ 1, versus normal, severity = 0). For each ET training strategy, we pooled test predictions across all backbone architectures and seeds. This differs deliberately from the convention used for the significance tests, where the eight per-reader Single-ET runs are averaged before pairing.

For the four-class scoring task, we focused on top-label calibration: Predicted maximum class probabilities were grouped into 10 uniform bins between 0 and 1. Within each bin, we computed the fraction of samples where the predicted top label matched its true label. Plotting the observed top-label accuracy vs. mean predicted top-label probability gave the reliability curves as in Supplementary Fig. 8. We also reported a vector ECE (ECEmulti) [36], defined as the average absolute gap between predicted and empirical class probabilities across bins and classes, and a multi-class Brier score (Briermulti) [37], defined as the mean squared error between predicted probability vectors and one-hot labels.

In addition, for the binary case, we collapsed the four classes into 'any lung disease' (severity ≥ 1) versus 'normal' (severity = 0). The probability of 'any disease' was defined as the sum of the predicted probabilities over the three diseased classes (severity 1-3). We computed the binary expected calibration error (ECE_binary) and binary Brier score (Brier_binary) in the same 10-bin framework. Because Single-ET is defined at the level of individual readers, it comprises eight reader-specific models per architecture and seed; each reader's metrics were computed on that reader's N = 162,078 pooled predictions and the eight values were then averaged, so the Single-ET row is comparable in sample size with the single-model strategies. Calibration results are summarized in Supplementary Table 5.

## 4.11 External validation datasets and inference protocol

To test whether gaze-guided training benefits generalize beyond the Emory cohort, we performed inference-only external validation on two independent, publicly available datasets. No model parameters were updated; all evaluations used Stage 2 checkpoints from the 13 training configurations, namely the no-gaze baseline, the eight reader-specific Single-ET models and the four panels (3Res-ET, 5Rad-ET, Mix5-ET, All8-ET), giving 910 models (13 x 7 architectures x 10 seeds (baseline + 8 per-reader Single-ET models + 3Res-ET + 5Rad-ET + Mix5-ET + All8-ET) × 7 architectures × 10 seeds) trained exclusively on the Emory data.

The first external dataset was the SIIM-FISABIO-RSNA 2021 COVID-19 chest-radiograph machine-learning challenge cohort [28]. The full challenge dataset contains 10,178 CXRs, split by the challenge organizers into a public training split (with labels) and a hidden test split (labels withheld). Because external validation requires ground-truth labels, we used the 6,054 studies (6,334 images) of the public training split as our external-validation cohort; Emory-trained models never saw any SIIM-FISABIO-RSNA data. Of the 6,054 studies, 3,373 are from male and 2,681 from female patients (ages not released in challenge Digital Imaging and Communications in Medicine (DICOM) headers) from two source cohorts, BIMCV-COVID19+ (Medical Imaging Databank of the Valencia Region, Spain) and MIDRC-RICORD (RSNA International COVID-19 Open Radiology Database), split across digital radiography (DX; 3,087 studies, 51.0%) and computed radiography (CR; 2,967 studies, 49.0%) modalities. Study-level labels were assigned by an international panel of 22 radiologists (19 staff subspecialty and 3 residents) into four mutually exclusive exam categories adapted from prior reporting guidelines [28, 40]: Negative for Pneumonia (1,676, 27.7%), Typical Appearance (2,855, 47.2%), Indeterminate Appearance (1,049, 17.3%) and Atypical Appearance (474, 7.8%). Inter-rater agreement was 86% (interquartile range (IQR) 64-91%, intraclass correlation coefficient (ICC) = 0.70, 95% confidence interval, CI 0.57-0.83) on 25 calibration cases [28]. For binary external validation we collapsed the four categories to 0 = Negative for Pneumonia and 1 = any pneumonia pattern (Typical/Indeterminate/Atypical; 4,378 abnormal, 72.3% prevalence). DICOM pixel arrays were decoded with rescale slope/intercept, windowed by WindowCenter/WindowWidth (or 1-99th percentile when absent), inverted for MONOCHROME1, resized to 1,024 pixels short side and saved as 8-bit JPEG (quality 95). The second external dataset was a subset of MIMIC-CXR [29] images included in MIMIC-Eye [14], which adds eye-tracking recordings from REFLACX [12]. From 3,689 MIMIC-Eye studies we retained 1,938 images with unambiguous CheXpert labels [3]: images labeled exclusively as 'No Finding' with no other positive pathology were classified as normal (1,157), while images positive for lung opacity, consolidation or pneumonia were classified as abnormal (781). Any other conditions were excluded.

Since severity scoring systems differ across institutions, we evaluated all models on a binary normal-versus-abnormal classification task. For each model, the four-class softmax outputs were collapsed post hoc as P(abnormal) = P(class 1) + P(class 2) + P(class 3). ROC-AUC and AU-PRC, which are threshold-independent, served as the primary metrics and are the basis of all comparative claims about generalization. We additionally report F1, sensitivity, and specificity at the Youden's J-optimal threshold computed per model on the external cohort; because this threshold is selected on the same data used to evaluate it, these threshold-dependent values are intended as descriptive summaries of the discrimination surface rather than as a clinically deployable operating point. Statistical comparisons between strategies followed the same paired Wilcoxon signed-rank framework described in Methods (Comparative statistics) but are reported without multiplicity correction, using 70 paired observations (7 architectures × 10 seeds) per strategy.

### 4.12 Implementation details

All models were implemented in Python using PyTorch (version 2.4.1+cu12) and trained on a Linux workstation with an Intel Xeon w5-2465X CPU (16 cores, 32 threads) and two NVIDIA RTX A4500 GPUs (20 GB each, driver 580.105.08, and CUDA 13.0). Each training run used a single GPU, where different architectures and seeds were scheduled concurrently across the two GPUs, without intra-model data parallelism. Training scripts, blackout-generation code, and configuration files were version-controlled to ensure exact experiment reproducibility.

All code and parameter settings needed to reproduce the experiments are available under reasonable request; see Code availability.

### 4.13 Ethics

The retrospective study was conducted under institutional review board oversight with waiver of informed consent. The eye-tracking recordings analyzed here were acquired under the same oversight in a prior AI-assisted observer-performance study.

### 4.14 Use of generative artificial intelligence

OpenAI and Anthropic models were used during manuscript preparation for language editing. The authors reviewed the source data, analysis code, numerical results, and final text and take responsibility for the work.

## 5 Data availability

The datasets employed in this study are available under reasonable request. Please contact Dr. Ali Adibi for any request regarding the data.

## 6 Code availability

The code employed in this study is available under reasonable request. Please contact Dr. Ali Adibi for any request regarding the code.

# 8 Acknowledgements

This work was supported by the National Science Foundation (NSF) under the grant titled Collaborative Research: SCH: Intelligent Radiology Through Human-Machine Cooperation, ICOL# 149725. We thank the radiologists and residents who contributed eye-tracking data.

# 9 Author contributions

All authors contributed to the study conception and design. Material preparation, data collection, and analysis were performed by all authors. The first draft of the manuscript was written by Nattakorn Kittisut, and all authors commented on previous versions of the manuscript. All authors read and approved the final manuscript.

# 10 Competing interests

The authors declare no competing interests.

# 11 Supplementary

**Supplementary Table 1 Overall performance of the per-model baseline and ET-based training strategies. Values are mean ± SD across architectures and seeds (in %). In each column, the best model performance is bolded.**

| Setting | F1 | ROC- AUC | AU-PRC |
|---|---|---|---|
| Baseline | 31.15 ± 6.68 | 64.66 ± 5.69 | 37.67 ± 5.68 |
| Single-ET | 33.02 ± 7.87 | 66.27 ± 5.82 | 38.73 ± 5.42 |
| 3Res-ET | 37.00 ± 11.44 | 69.37 ± 9.30 | 42.21 ± 8.77 |
| 5Rad-ET | 36.54 ± 12.37 | 68.99 ± 10.60 | 41.97 ± 9.68 |
| Mix5-ET | 36.48 ± 11.34 | 68.78 ± 9.75 | 41.38 ± 8.98 |
| **All8-ET** | **38.62 ± 12.01** | **70.56 ± 9.60** | **43.87 ± 9.32** |

**Supplementary Table 2 Error profile of the per-model baseline and ET-based strategies. Values are mean ± SD across architectures and seeds (in %). All four multi-reader strategies improve recall and macro specificity relative to the baseline, with All8-ET largest on each; on weighted specificity only All8-ET reaches $p_{adj} < 0.05$. In each column, the best model performance is bolded.**

| Setting | Recall | Precision | $\text{Specificity}_{\text{macro}}$ | $\text{Specificity}_{\text{weighted}}$ |
|---|---|---|---|---|
| Baseline | 36.65 ± 5.94 | 35.82 ± 7.24 | 78.61 ± 1.75 | 76.45 ± 4.31 |
| Single-ET | 37.61 ± 5.51 | 35.61 ± 9.11 | 78.78 ± 1.70 | 76.91 ± 3.30 |
| 3Res-ET | 40.61 ± 8.30 | 39.76 ± 12.15 | 80.01 ± 2.81 | 78.28 ± 4.64 |
| 5Rad-ET | 40.19 ± 9.32 | 39.00 ± 14.06 | 79.84 ± 3.03 | 77.74 ± 4.81 |
| Mix5-ET | 40.03 ± 8.54 | 39.04 ± 13.17 | 79.68 ± 2.81 | 77.50 ± 4.94 |
| **All8-ET** | **42.46 ± 9.25** | **41.51 ± 12.75** | **80.31 ± 2.93** | **78.59 ± 4.64** |

**Supplementary Table 3 Per-architecture comparison of All8-ET against Single-ET, with Baseline shown for reference. Single-ET levels, the subtrahend of the delta column, are in Supplementary Table 4. For each architecture and evaluation metric, we report the mean performance (μ ± σ, in %) across 10 random seeds for the per-model Baseline and the All8-ET strategy, and the paired delta Δ (All8-ET − Single-ET, in percentage points), paired Wilcoxon p values, and Cohen's $d_z$. Significant at uncorrected p < 0.05 marked by "*".**

| Model | Metric | Baseline | All8-ET | Δ (All8−Single) | *p* | $d_z$ |
|---|---|---|---|---|---|---|
| DenseNet-121 | F1 | 35.17 ± 3.55 | 45.56 ± 2.15 | +7.74 ± 3.91 | 0.0020* | 1.98 |
| | Recall | 39.48 ± 3.67 | 46.59 ± 2.68 | +5.79 ± 4.17 | 0.0059* | 1.39 |
| | Precision | 37.90 ± 3.83 | 48.58 ± 4.24 | +7.11 ± 5.63 | 0.0020* | 1.26 |
| | ROC-AUC | 66.42 ± 3.41 | 76.22 ± 3.59 | +5.25 ± 4.40 | 0.0059* | 1.19 |
| | AU-PRC | 38.54 ± 3.67 | 49.19 ± 4.40 | +6.69 ± 5.53 | 0.0059* | 1.21 |
| | $Spec_{macro}$ | 79.28 ± 0.92 | 81.91 ± 1.16 | +1.96 ± 1.51 | 0.0098* | 1.30 |
| | $Spec_{weighted}$ | 77.92 ± 2.03 | 79.61 ± 1.35 | +0.86 ± 2.06 | 0.1934 | 0.42 |
| EfficientNet-B0 | F1 | 26.46 ± 5.41 | 39.41 ± 2.36 | +2.39 ± 4.57 | 0.1309 | 0.52 |
| | Recall | 31.31 ± 2.82 | 42.39 ± 2.18 | +3.63 ± 3.86 | 0.0137* | 0.94 |
| | Precision | 36.31 ± 5.16 | 42.51 ± 3.46 | +1.87 ± 6.07 | 0.2754 | 0.31 |
| | ROC-AUC | 62.63 ± 5.47 | 71.00 ± 2.68 | +3.17 ± 5.28 | 0.0840 | 0.60 |
| | AU-PRC | 35.36 ± 4.90 | 43.75 ± 3.23 | +2.91 ± 5.90 | 0.1934 | 0.49 |
| | $Spec_{macro}$ | 77.11 ± 0.92 | 79.94 ± 0.97 | +0.53 ± 1.35 | 0.2754 | 0.39 |
| | $Spec_{weighted}$ | 70.89 ± 3.12 | 78.29 ± 1.87 | +1.64 ± 3.00 | 0.1309 | 0.55 |
| EfficientNet-B1 | F1 | 33.07 ± 8.00 | 40.83 ± 3.42 | +3.71 ± 3.58 | 0.0195* | 1.04 |
| | Recall | 39.01 ± 5.56 | 44.35 ± 3.27 | +3.77 ± 3.94 | 0.0273* | 0.96 |
| | Precision | 40.54 ± 9.48 | 41.90 ± 2.29 | +3.03 ± 2.47 | 0.0059* | 1.23 |
| | ROC-AUC | 68.11 ± 4.01 | 71.59 ± 2.57 | +3.54 ± 1.91 | 0.0039* | 1.86 |
| | AU-PRC | 40.90 ± 3.95 | 43.93 ± 3.74 | +3.16 ± 2.54 | 0.0098* | 1.24 |
| | $Spec_{macro}$ | 79.50 ± 1.81 | 80.55 ± 0.94 | +0.96 ± 1.11 | 0.0371* | 0.86 |
| | $Spec_{weighted}$ | 76.39 ± 3.73 | 80.21 ± 1.51 | +1.62 ± 2.17 | 0.0371* | 0.75 |
| ResNet-18 | F1 | 32.45 ± 4.84 | 47.24 ± 2.51 | +13.41 ± 5.29 | 0.0020* | 2.54 |
| | Recall | 37.22 ± 4.67 | 49.33 ± 2.72 | +10.58 ± 4.53 | 0.0020* | 2.34 |
| | Precision | 37.63 ± 6.41 | 48.88 ± 2.01 | +10.82 ± 2.90 | 0.0020* | 3.73 |
| | ROC-AUC | 65.02 ± 3.58 | 77.42 ± 1.99 | +9.34 ± 4.10 | 0.0020* | 2.28 |
| | AU-PRC | 38.16 ± 3.44 | 50.91 ± 2.47 | +11.52 ± 3.65 | 0.0020* | 3.16 |
| | $Spec_{macro}$ | 78.86 ± 1.59 | 82.52 ± 1.00 | +3.79 ± 1.64 | 0.0020* | 2.32 |
| | $Spec_{weighted}$ | 77.78 ± 4.41 | 80.96 ± 1.54 | +2.83 ± 2.12 | 0.0020* | 1.33 |
| ResNet-50 | F1 | 28.40 ± 3.93 | 42.11 ± 5.50 | +7.86 ± 5.52 | 0.0020* | 1.42 |
| | Recall | 38.32 ± 2.94 | 45.45 ± 4.48 | +6.47 ± 4.81 | 0.0039* | 1.35 |
| | Precision | 31.04 ± 4.70 | 47.97 ± 3.26 | +11.37 ± 4.26 | 0.0020* | 2.67 |
| | ROC-AUC | 64.59 ± 3.21 | 74.68 ± 2.66 | +7.71 ± 3.64 | 0.0020* | 2.12 |
| | AU-PRC | 36.86 ± 2.85 | 47.53 ± 3.65 | +8.82 ± 4.49 | 0.0020* | 1.96 |
| | $Spec_{macro}$ | 78.83 ± 0.95 | 81.35 ± 1.54 | +2.54 ± 1.72 | 0.0039* | 1.47 |
| | $Spec_{weighted}$ | 78.88 ± 3.61 | 80.77 ± 2.95 | +2.22 ± 3.23 | 0.1055 | 0.69 |
| Swin | F1 | 38.04 ± 5.63 | 38.09 ± 18.94 | +4.43 ± 15.87 | 0.3750 | 0.28 |
| | Recall | 42.64 ± 4.44 | 42.80 ± 15.07 | +4.50 ± 12.21 | 0.3750 | 0.37 |
| | Precision | 40.20 ± 4.48 | 39.23 ± 19.68 | +2.51 ± 16.43 | 0.6953 | 0.15 |
| | ROC-AUC | 70.24 ± 3.62 | 69.46 ± 14.92 | +3.13 ± 11.75 | 0.4922 | 0.27 |
| | AU-PRC | 44.55 ± 4.49 | 43.60 ± 14.81 | +3.97 ± 11.96 | 0.3750 | 0.33 |
| | $Spec_{macro}$ | 80.18 ± 1.58 | 80.57 ± 4.68 | +1.47 ± 4.06 | 0.2754 | 0.36 |
| | $Spec_{weighted}$ | 80.04 ± 1.82 | 80.25 ± 6.27 | +2.96 ± 5.95 | 0.1309 | 0.50 |
| ViT | F1 | 24.47 ± 2.48 | 17.12 ± 4.38 | -0.32 ± 4.44 | 0.7695 | -0.07 |
| | Recall | 28.60 ± 2.43 | 26.28 ± 1.42 | -0.84 ± 1.39 | 0.0840 | -0.61 |

| Model | Metric | Baseline | All8-ET | Δ (All8−Single) | *p* | $d_z$ |
|---|---|---|---|---|---|---|
| | Precision | 27.14 ± 4.82 | 21.51 ± 13.98 | +4.62 ± 12.45 | 0.9219 | 0.37 |
| | ROC-AUC | 55.63 ± 2.78 | 53.55 ± 3.99 | -2.08 ± 4.93 | 0.2324 | -0.42 |
| | AU-PRC | 29.31 ± 2.17 | 28.15 ± 2.75 | -1.11 ± 3.61 | 0.4922 | -0.31 |
| | $Spec_{macro}$ | 76.48 ± 0.91 | 75.34 ± 0.45 | -0.50 ± 0.68 | 0.0645 | -0.73 |
| | $Spec_{weighted}$ | 73.24 ± 2.85 | 70.00 ± 2.47 | -0.38 ± 2.52 | 0.6953 | -0.15 |

**Supplementary Table 4 Per-architecture performance with and without ET supervision. For each backbone and evaluation metric, we report mean μ ± σ performance (in %) across 10 random seeds for the per-model Baseline and five ET strategies (Single-ET, 3Res-ET, Mix5-ET, 5Rad-ET, All8-ET).**

| Model | Metric | Baseline | Single-ET | 3Res-ET | Mix5-ET | 5Rad-ET | All8-ET |
|---|---|---|---|---|---|---|---|
| DenseNet-121 | F1 | 35.17 ± 3.55 | 37.82 ± 2.20 | 43.59 ± 6.55 | 44.64 ± 4.85 | 43.43 ± 4.22 | 45.56 ± 2.15 |
| | Recall | 39.48 ± 3.67 | 40.80 ± 2.01 | 45.25 ± 5.05 | 45.14 ± 4.20 | 44.08 ± 4.10 | 46.59 ± 2.68 |
| | Precision | 37.90 ± 3.83 | 41.47 ± 2.94 | 47.88 ± 6.48 | 50.66 ± 5.84 | 48.35 ± 6.45 | 48.58 ± 4.24 |
| | ROC-AUC | 66.42 ± 3.41 | 70.97 ± 2.31 | 77.05 ± 4.64 | 77.15 ± 4.09 | 75.86 ± 4.83 | 76.22 ± 3.59 |
| | AU-PRC | 38.54 ± 3.67 | 42.50 ± 2.64 | 49.40 ± 5.67 | 49.23 ± 5.12 | 47.91 ± 5.71 | 49.19 ± 4.40 |
| | $Spec_{macro}$ | 79.28 ± 0.92 | 79.95 ± 0.73 | 82.13 ± 2.24 | 82.15 ± 1.67 | 81.45 ± 1.60 | 81.91 ± 1.16 |
| | $Spec_{weighted}$ | 77.92 ± 2.03 | 78.75 ± 1.17 | 80.05 ± 2.67 | 80.58 ± 1.85 | 79.79 ± 1.89 | 79.61 ± 1.35 |
| EfficientNet-B0 | F1 | 26.46 ± 5.41 | 37.03 ± 3.67 | 39.82 ± 6.20 | 37.72 ± 3.31 | 36.96 ± 4.09 | 39.41 ± 2.36 |
| | Recall | 31.31 ± 2.82 | 38.75 ± 2.82 | 41.28 ± 4.50 | 40.20 ± 3.25 | 39.99 ± 3.55 | 42.39 ± 2.18 |
| | Precision | 36.31 ± 5.16 | 40.64 ± 4.34 | 42.02 ± 5.24 | 40.54 ± 4.42 | 41.26 ± 3.00 | 42.51 ± 3.46 |
| | ROC-AUC | 62.63 ± 5.47 | 67.83 ± 4.05 | 69.80 ± 4.03 | 69.29 ± 3.14 | 69.48 ± 2.55 | 71.00 ± 2.68 |
| | AU-PRC | 35.36 ± 4.90 | 40.84 ± 4.19 | 42.44 ± 5.39 | 41.07 ± 3.97 | 41.61 ± 2.55 | 43.75 ± 3.23 |
| | $Spec_{macro}$ | 77.11 ± 0.92 | 79.42 ± 1.05 | 79.86 ± 1.77 | 79.20 ± 1.03 | 79.25 ± 1.04 | 79.94 ± 0.97 |
| | $Spec_{weighted}$ | 70.89 ± 3.12 | 76.65 ± 1.73 | 78.51 ± 1.62 | 78.14 ± 1.56 | 78.35 ± 1.25 | 78.29 ± 1.87 |
| EfficientNet-B1 | F1 | 33.07 ± 8.00 | 37.12 ± 4.48 | 38.83 ± 5.25 | 36.23 ± 4.79 | 39.07 ± 5.68 | 40.83 ± 3.42 |
| | Recall | 39.01 ± 5.56 | 40.58 ± 3.03 | 42.25 ± 3.71 | 40.64 ± 3.64 | 41.61 ± 4.48 | 44.35 ± 3.27 |
| | Precision | 40.54 ± 9.48 | 38.87 ± 3.49 | 40.07 ± 4.40 | 37.85 ± 3.64 | 40.49 ± 4.59 | 41.90 ± 2.29 |
| | ROC-AUC | 68.11 ± 4.01 | 68.05 ± 3.00 | 69.05 ± 4.01 | 67.61 ± 3.34 | 69.09 ± 3.78 | 71.59 ± 2.57 |
| | AU-PRC | 40.90 ± 3.95 | 40.77 ± 3.26 | 41.47 ± 4.55 | 39.26 ± 3.80 | 41.28 ± 4.75 | 43.93 ± 3.74 |
| | $Spec_{macro}$ | 79.50 ± 1.81 | 79.59 ± 1.15 | 79.99 ± 1.45 | 79.11 ± 1.47 | 79.81 ± 1.76 | 80.55 ± 0.94 |
| | $Spec_{weighted}$ | 76.39 ± 3.73 | 78.59 ± 1.20 | 79.27 ± 1.69 | 78.75 ± 2.11 | 78.49 ± 2.37 | 80.21 ± 1.51 |
| ResNet-18 | F1 | 32.45 ± 4.84 | 33.83 ± 3.81 | 44.05 ± 3.11 | 44.11 ± 5.09 | 45.08 ± 3.53 | 47.24 ± 2.51 |
| | Recall | 37.22 ± 4.67 | 38.75 ± 2.92 | 45.91 ± 3.94 | 44.76 ± 4.49 | 46.66 ± 2.84 | 49.33 ± 2.72 |
| | Precision | 37.63 ± 6.41 | 38.06 ± 2.91 | 47.06 ± 3.23 | 47.22 ± 4.39 | 49.26 ± 3.85 | 48.88 ± 2.01 |
| | ROC-AUC | 65.02 ± 3.58 | 68.08 ± 3.45 | 75.41 ± 2.74 | 75.02 ± 3.48 | 77.39 ± 3.28 | 77.42 ± 1.99 |
| | AU-PRC | 38.16 ± 3.44 | 39.39 ± 3.03 | 47.74 ± 3.31 | 47.81 ± 4.22 | 49.98 ± 4.00 | 50.91 ± 2.47 |
| | $Spec_{macro}$ | 78.86 ± 1.59 | 78.72 ± 0.90 | 81.92 ± 1.08 | 81.45 ± 1.83 | 82.06 ± 1.26 | 82.52 ± 1.00 |
| | $Spec_{weighted}$ | 77.78 ± 4.41 | 78.14 ± 1.32 | 80.83 ± 1.82 | 79.28 ± 2.74 | 80.37 ± 1.47 | 80.96 ± 1.54 |
| ResNet-50 | F1 | 28.40 ± 3.93 | 34.24 ± 1.96 | 40.30 ± 3.87 | 43.25 ± 3.57 | 42.41 ± 5.97 | 42.11 ± 5.50 |
| | Recall | 38.32 ± 2.94 | 38.98 ± 1.58 | 42.46 ± 3.22 | 45.78 ± 3.47 | 45.16 ± 5.34 | 45.45 ± 4.48 |
| | Precision | 31.04 ± 4.70 | 36.60 ± 2.56 | 44.13 ± 4.09 | 46.28 ± 2.93 | 45.74 ± 2.10 | 47.97 ± 3.26 |
| | ROC-AUC | 64.59 ± 3.21 | 66.97 ± 1.86 | 72.56 ± 3.45 | 74.91 ± 2.37 | 74.74 ± 2.36 | 74.68 ± 2.66 |
| | AU-PRC | 36.86 ± 2.85 | 38.72 ± 1.89 | 44.06 ± 3.96 | 46.38 ± 3.08 | 46.46 ± 3.14 | 47.53 ± 3.65 |
| | $Spec_{macro}$ | 78.83 ± 0.95 | 78.81 ± 0.74 | 80.58 ± 1.32 | 81.33 ± 1.23 | 81.44 ± 1.68 | 81.35 ± 1.54 |
| | $Spec_{weighted}$ | 78.88 ± 3.61 | 78.55 ± 1.03 | 79.74 ± 1.68 | 80.34 ± 1.48 | 80.53 ± 1.31 | 80.77 ± 2.95 |
| Swin | F1 | 38.04 ± 5.63 | 33.67 ± 8.31 | 36.19 ± 15.51 | 32.70 ± 15.25 | 31.29 ± 20.71 | 38.09 ± 18.94 |
| | Recall | 42.64 ± 4.44 | 38.30 ± 7.04 | 40.90 ± 12.09 | 37.35 ± 13.29 | 37.74 ± 16.13 | 42.80 ± 15.07 |
| | Precision | 40.20 ± 4.48 | 36.73 ± 8.60 | 39.44 ± 15.61 | 34.64 ± 16.85 | 31.13 ± 21.97 | 39.23 ± 19.68 |
| | ROC-AUC | 70.24 ± 3.62 | 66.33 ± 6.92 | 68.77 ± 11.92 | 64.03 ± 13.82 | 64.49 ± 16.03 | 69.46 ± 14.92 |
| | AU-PRC | 44.55 ± 4.49 | 39.63 ± 6.97 | 42.30 ± 12.40 | 37.64 ± 13.07 | 39.18 ± 15.59 | 43.60 ± 14.81 |
| | $Spec_{macro}$ | 80.18 ± 1.58 | 79.09 ± 2.25 | 80.11 ± 3.82 | 78.86 ± 4.05 | 79.21 ± 5.25 | 80.57 ± 4.68 |
| | $Spec_{weighted}$ | 80.04 ± 1.82 | 77.29 ± 3.24 | 78.70 ± 5.60 | 76.34 ± 6.98 | 76.51 ± 8.16 | 80.25 ± 6.27 |
| ViT | F1 | 24.47 ± 2.48 | 17.44 ± 3.87 | 16.25 ± 4.69 | 16.71 ± 3.90 | 17.57 ± 4.43 | 17.12 ± 4.38 |
| | Recall | 28.60 ± 2.43 | 27.12 ± 1.45 | 26.18 ± 1.59 | 26.36 ± 1.50 | 26.13 ± 1.01 | 26.28 ± 1.42 |
| | Precision | 27.14 ± 4.82 | 16.89 ± 5.61 | 17.72 ± 8.49 | 16.06 ± 7.47 | 16.80 ± 7.25 | 21.51 ± 13.98 |
| | ROC-AUC | 55.63 ± 2.78 | 55.63 ± 2.28 | 52.92 ± 5.41 | 53.47 ± 5.65 | 51.90 ± 5.38 | 53.55 ± 3.99 |
| | AU-PRC | 29.31 ± 2.17 | 29.27 ± 1.52 | 28.08 ± 3.18 | 28.28 ± 3.15 | 27.37 ± 2.90 | 28.15 ± 2.75 |
| | $Spec_{macro}$ | 76.48 ± 0.91 | 75.84 ± 0.71 | 75.46 ± 0.71 | 75.67 ± 0.74 | 75.67 ± 0.57 | 75.34 ± 0.45 |

| Model | Metric | Baseline | Single-ET | 3Res-ET | Mix5-ET | 5Rad-ET | All8-ET |
|---|---|---|---|---|---|---|---|
| | $Spec_{weighted}$ | 73.24 ± 2.85 | 70.38 ± 2.34 | 70.86 ± 6.29 | 69.06 ± 3.54 | 70.16 ± 2.89 | 70.00 ± 2.47 |

**Supplementary Table 5 Calibration of four-class severity and binary (any-disease vs. normal) predictions across training strategies, pooled across the seven architectures and 10 seeds. The no-gaze baseline row was produced by running inference on its 70 saved checkpoints, so every strategy is referenced to it. ECE_binary and Brier_binary are the expected calibration error and Brier score for the any-disease versus normal collapse; ECE_multi is the vector expected calibration error and Brier_multi the multi-class Brier score (mean squared error between predicted probability vectors and one-hot labels). For Single-ET, the metrics are computed separately for each of the eight reader-specific models, each on N = 162,078 pooled predictions, and the eight values are averaged, so the row stays comparable with the single-model strategies.**

| Strategy | *N* | $ECE_{binary}$ | $Brier_{binary}$ | $ECE_{multi}$ | Briermult i |
|---|---|---|---|---|---|
| Baseline | 162 078 | 0.1417 | 0.1027 | 0.3011 | 0.1815 |
| Single-ET | 162 078 | 0.1158 | 0.0958 | 0.2918 | 0.1767 |
| 3Res-ET | 162 078 | 0.0795 | 0.0868 | 0.2745 | 0.1701 |
| Mix5-ET | 162 078 | 0.0781 | 0.0873 | 0.2882 | 0.1728 |
| 5Rad-ET | 162 078 | 0.0693 | 0.0847 | 0.2939 | 0.1713 |
| All8-ET | 162 078 | 0.0733 | 0.0848 | 0.2832 | 0.1716 |

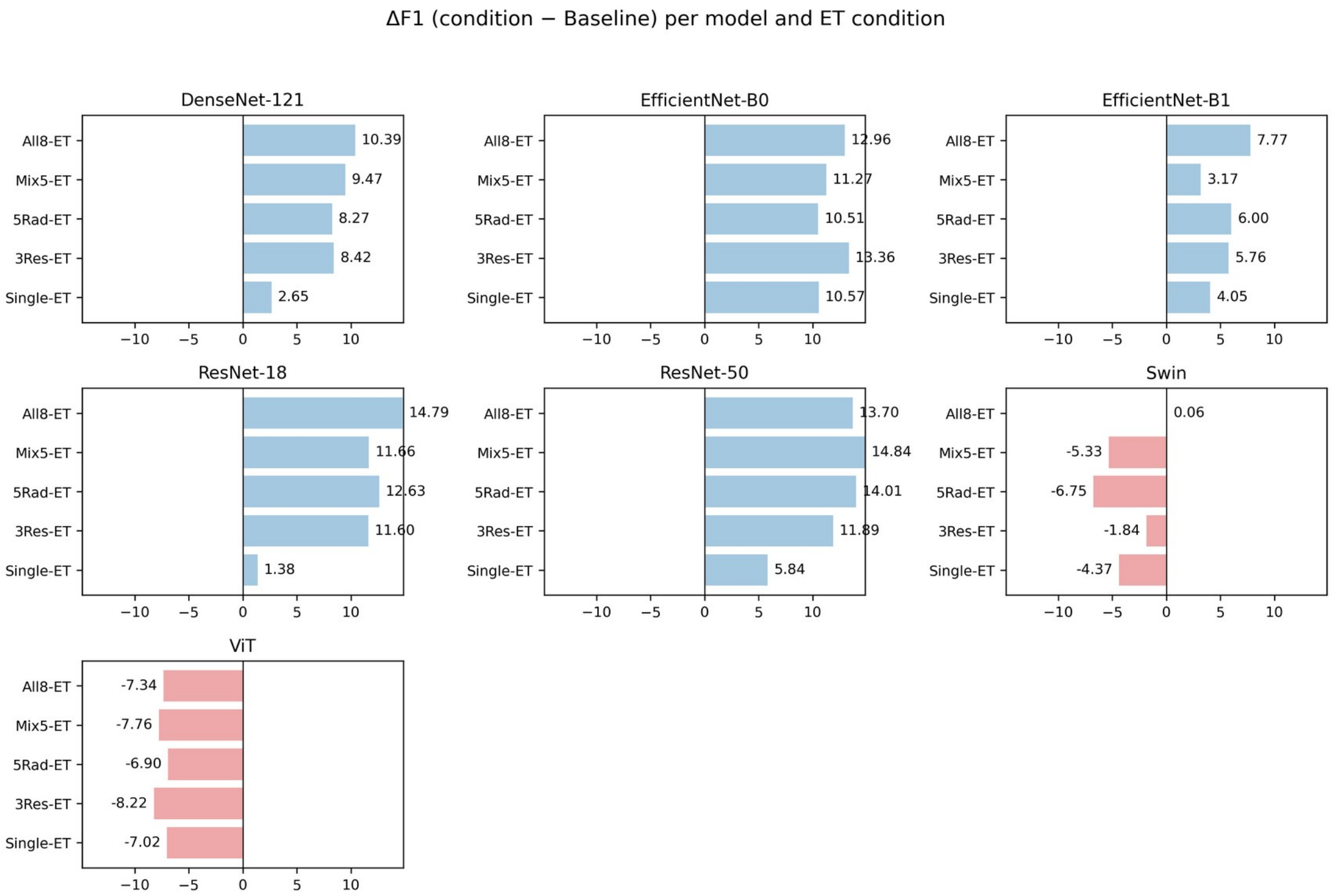

Supplementary Fig. 1 Architecture-specific F1 gains for each ET strategy. Change in macro F1 (pp) relative to baseline for each backbone (rows) and ET strategy (bars). CNN architectures (DenseNet-121, EfficientNet-B0/B1, ResNet-18/50) show large, consistent improvements with All8-ET, whereas Swin is essentially unchanged under All8-ET and loses ground under the subset strategies, and ViT degrades throughout.

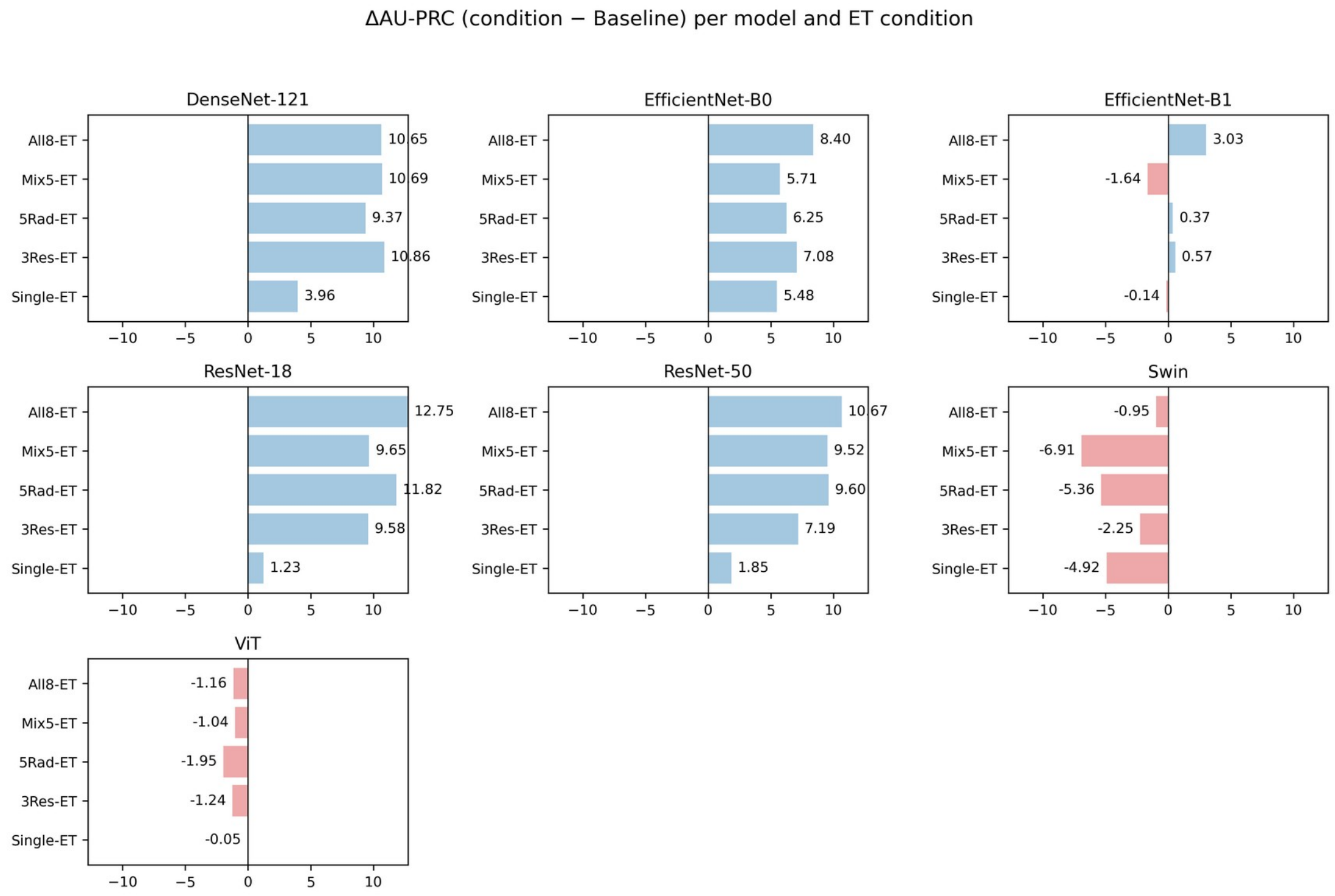

Supplementary Fig. 2 Architecture-specific AU-PRC gains. Change in AU-PRC (pp) relative to baseline for each backbone and ET strategy. Patterns mirror those in F1, with the largest gains for CNNs under All8-ET.

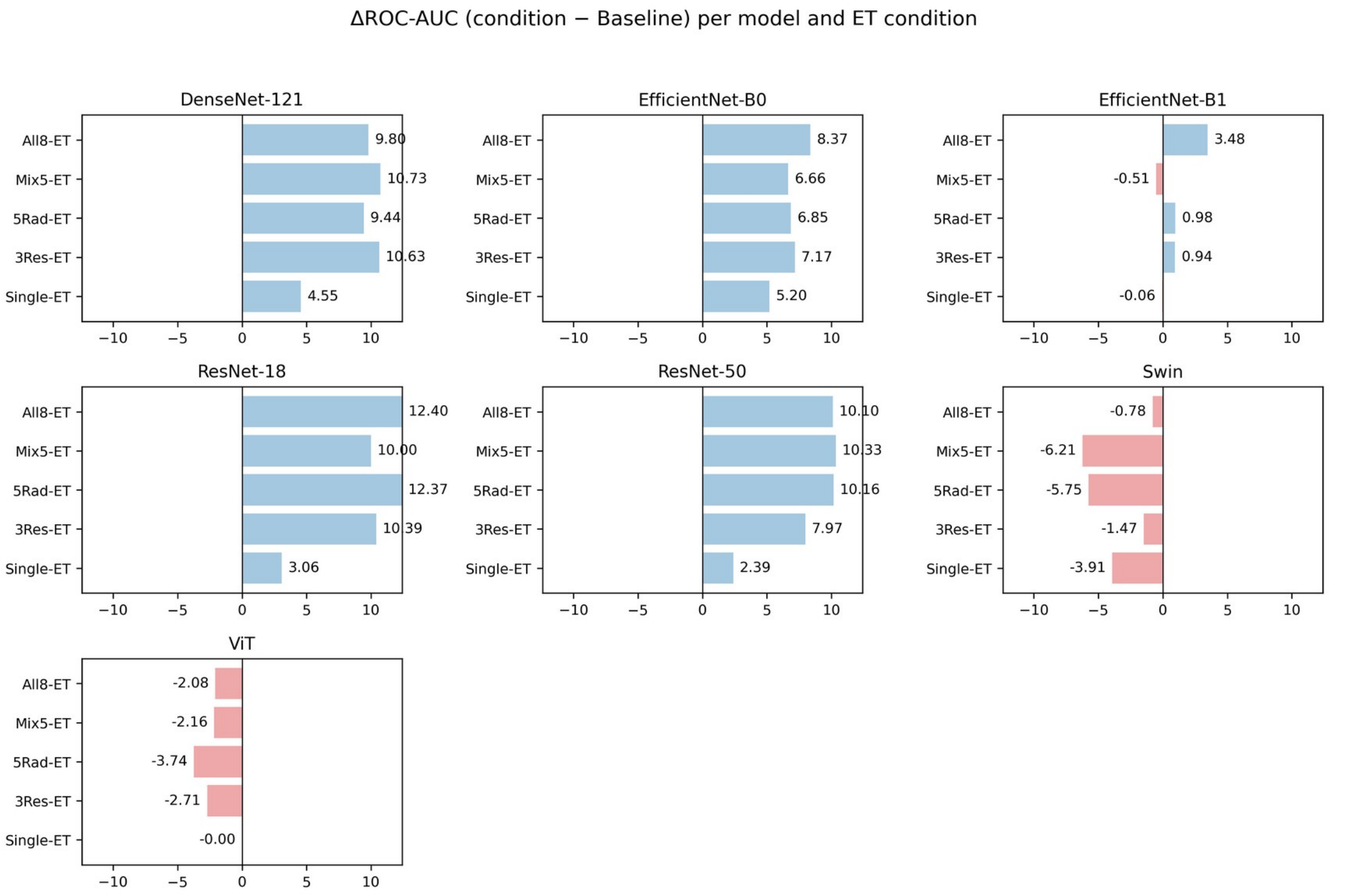

Supplementary Fig. 3 Architecture-specific ROC-AUC gains. Change in ROC-AUC (pp) relative to baseline for each backbone and ET strategy.

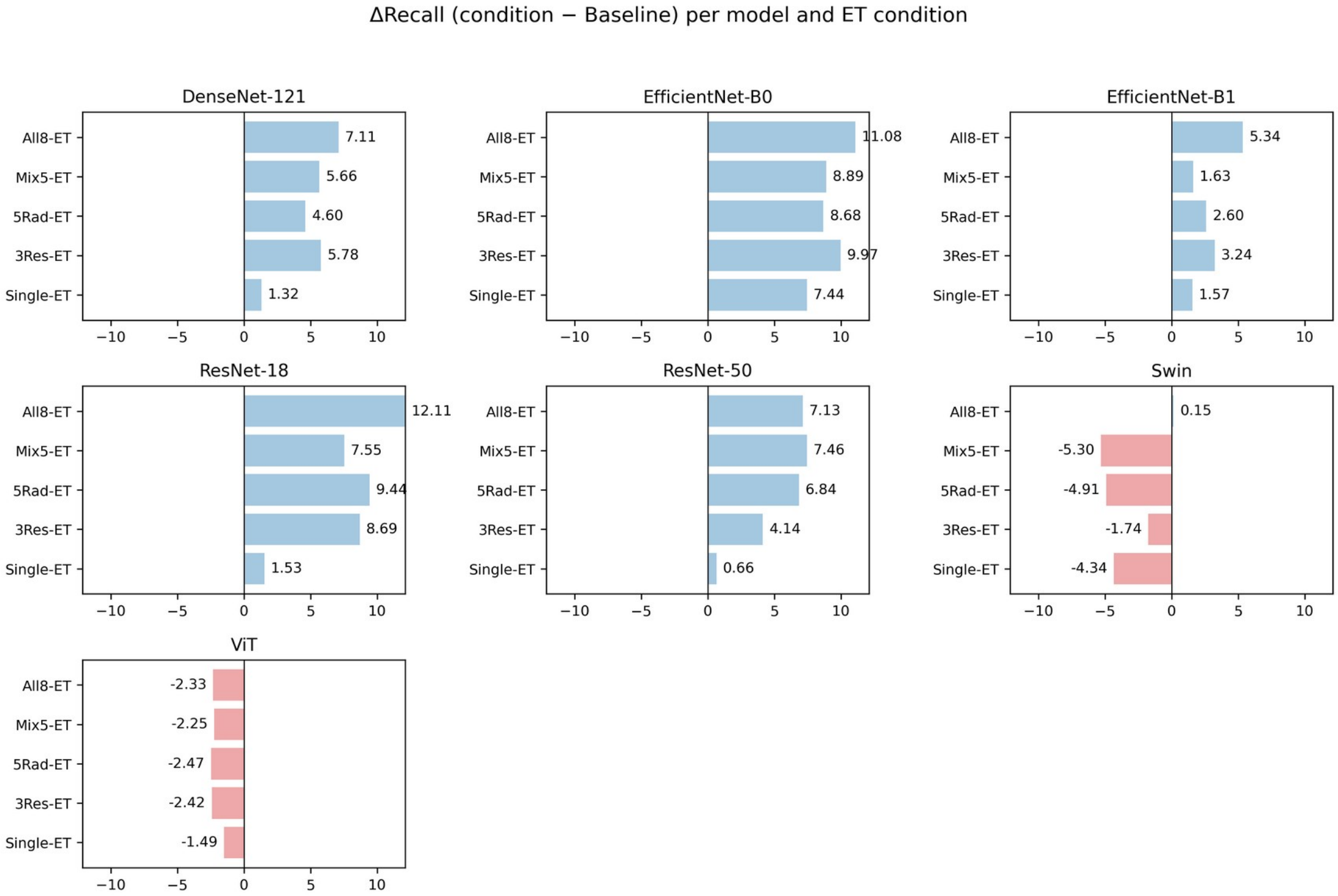

Supplementary Fig. 4 Architecture-specific recall gains. Change in recall (pp) relative to baseline for each backbone and ET strategy.

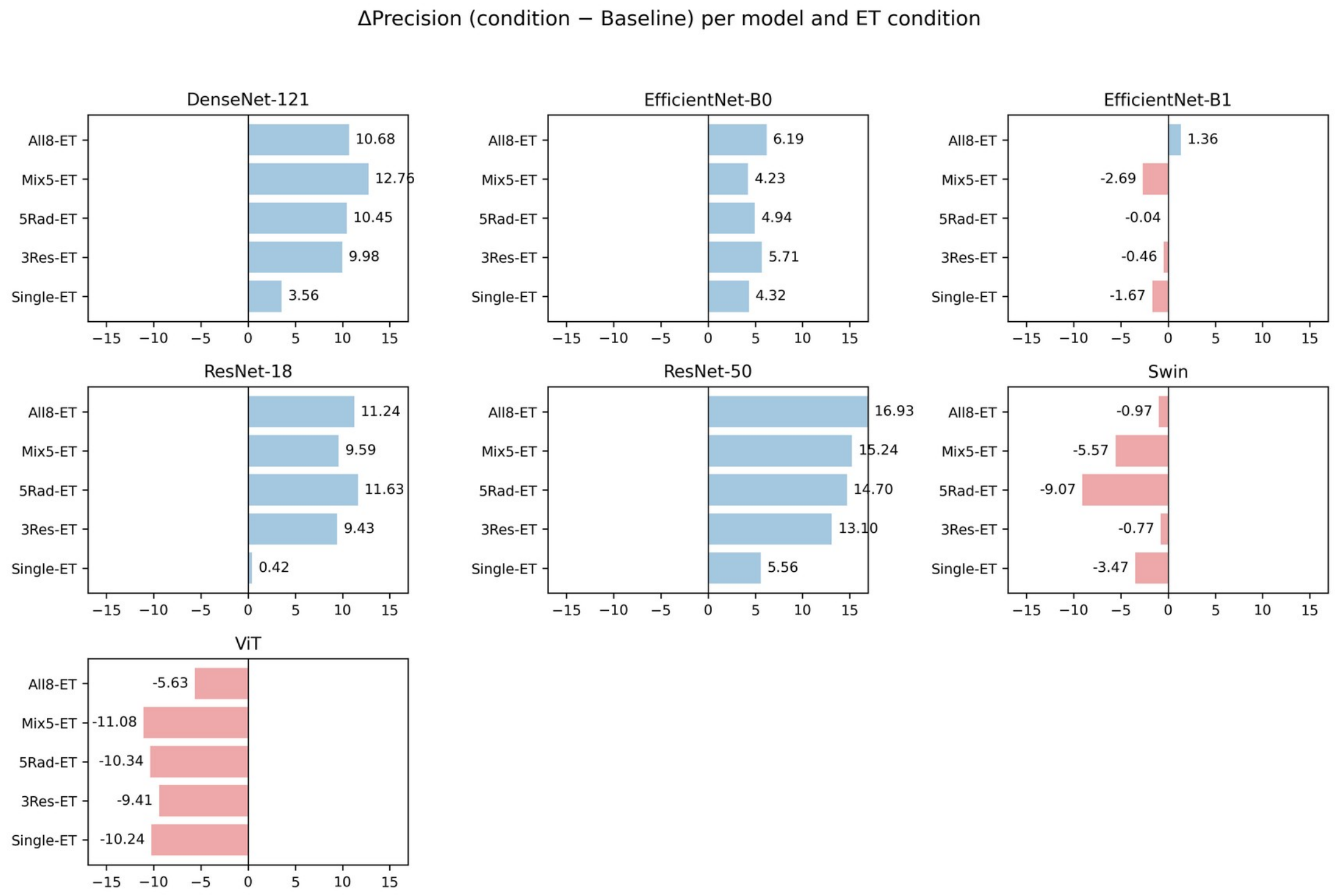

Supplementary Fig. 5 Architecture-specific precision changes. Change in precision (pp) relative to baseline for each backbone and ET strategy, highlighting precision-recall trade-offs for some Single-ET models and transformers.

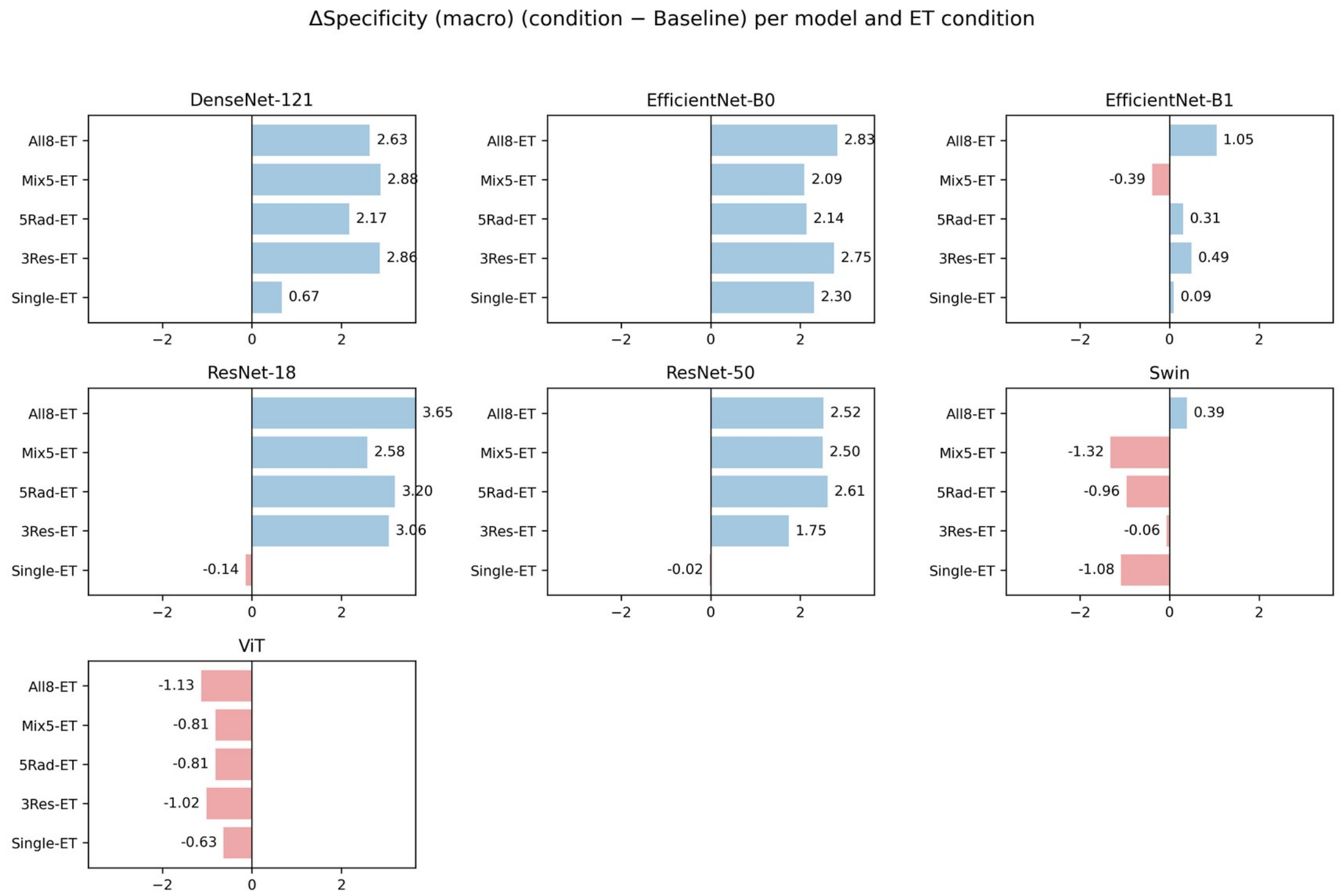

Supplementary Fig. 6 Architecture-specific changes in macro specificity. Change in macro specificity (pp) relative to baseline for each backbone and ET strategy.

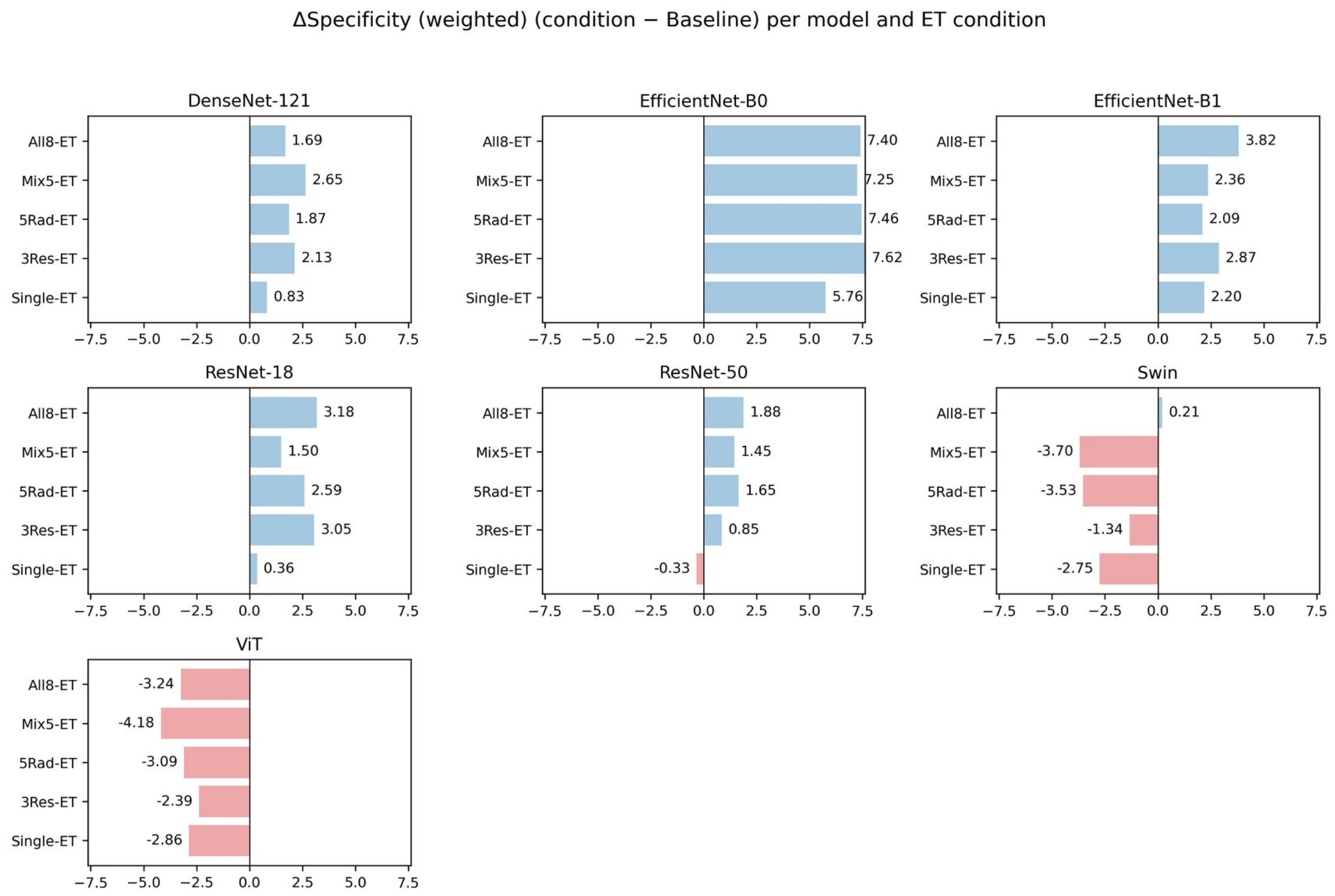

Supplementary Fig. 7 Architecture-specific changes in weighted specificity. Change in weighted specificity (pp) relative to baseline for each backbone and ET strategy.

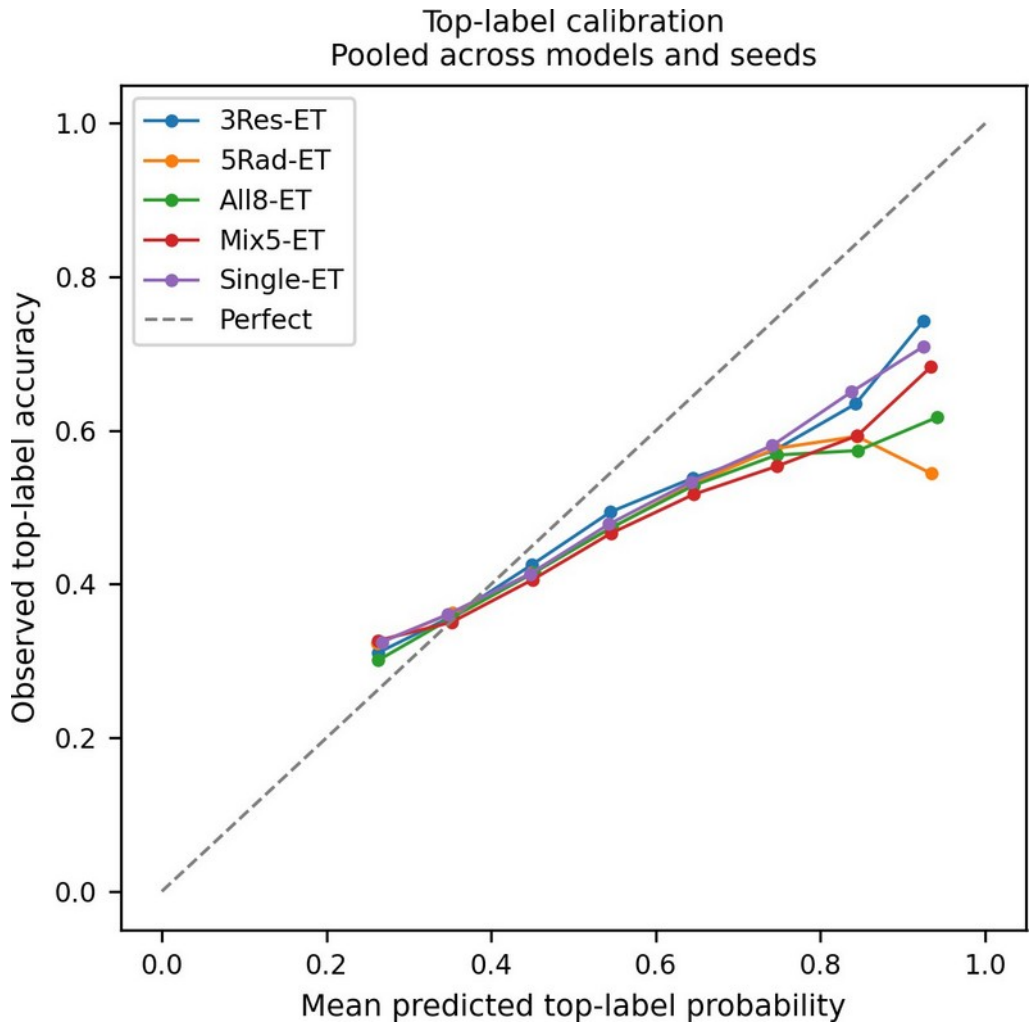

Supplementary Fig. 8 Top-label calibration of four-class severity predictions across gaze-guided training strategies, pooled across architectures and seeds. All strategies show broadly similar reliability. Three of the four multi-reader approaches (3Res-ET, Mix5-ET and All8-ET) achieve slightly lower miscalibration than Single-ET, whereas 5Rad-ET is marginally higher on this four-class metric (vector expected calibration error 0.294 versus 0.292). The dashed line indicates perfect calibration.

**Supplementary Table 6 Per-architecture external validation performance on the SIIM-FISABIO-RSNA 2021 COVID-19 cohort (N = 6,054 studies, 6,334 images; 55.7% male; binary label: 0 = Negative for Pneumonia, 1 = any of Typical / Indeterminate / Atypical Appearance). For each backbone architecture and metric, we report mean ± SD (%) across 10 random seeds. All metrics except ROC-AUC and AU-PRC are computed at the Youden's J-optimal threshold. Single-ET is the pooled average over the eight individual-reader strategies (80 models per cell). ViT entries on SIIM-FISABIO-RSNA reflect near-degenerate binary predictions at the Youden-optimal threshold, in opposite directions for the gaze-trained and baseline rows: the gaze-trained ViT models shift toward high sensitivity and low specificity (sensitivity 57-87%, specificity 14-46% across the five gaze strategies), whereas the ViT baseline converges the other way (mean sensitivity 26% and specificity 80%); the discriminative summary for ViT is therefore the threshold-independent ROC-AUC, which is 0.51 to 0.53 for every strategy including the baseline, not F1 or sensitivity/specificity, and gaze guidance did not rescue this.**

| Model | Metric | Baseline | Single-ET | 3Res-ET | Mix5-ET | 5Rad-ET | All8-ET |
|---|---|---|---|---|---|---|---|
| DenseNet-121 | ROC-AUC | 69.4 ± 6.4 | 73.1 ± 4.4 | 77.8 ± 4.3 | 78.7 ± 3.9 | 77.1 ± 4.5 | 79.7 ± 2.7 |
| | AU-PRC | 85.6 ± 3.4 | 87.5 ± 2.5 | 90.1 ± 2.3 | 90.6 ± 2.1 | 89.6 ± 2.8 | 91.1 ± 1.4 |
| | F1 | 65.0 ± 12.8 | 71.3 ± 4.3 | 75.0 ± 4.8 | 74.9 ± 5.9 | 74.1 ± 3.8 | 77.1 ± 3.9 |
| | Sensitivity | 54.1 ± 14.5 | 61.0 ± 5.7 | 65.4 ± 6.6 | 64.9 ± 8.0 | 64.0 ± 4.7 | 68.2 ± 5.7 |
| | Specificity | 74.7 ± 8.6 | 74.1 ± 5.0 | 76.9 ± 3.9 | 79.2 ± 4.0 | 77.2 ± 3.7 | 77.3 ± 2.7 |
| ResNet-18 | ROC-AUC | 74.6 ± 4.3 | 73.3 ± 6.3 | 79.0 ± 4.6 | 80.1 ± 2.4 | 81.2 ± 2.6 | 81.9 ± 2.9 |
| | AU-PRC | 87.8 ± 2.2 | 87.9 ± 3.6 | 90.7 ± 2.1 | 90.9 ± 1.4 | 91.6 ± 1.3 | 92.0 ± 1.2 |
| | F1 | 75.1 ± 4.1 | 71.2 ± 5.6 | 76.7 ± 5.8 | 78.5 ± 2.7 | 79.8 ± 2.9 | 79.8 ± 4.3 |
| | Sensitivity | 67.2 ± 5.8 | 61.0 ± 8.2 | 68.2 ± 8.2 | 70.9 ± 4.1 | 72.6 ± 4.7 | 72.4 ± 6.4 |
| | Specificity | 69.4 ± 4.3 | 74.2 ± 10.7 | 76.3 ± 2.8 | 74.9 ± 3.3 | 75.6 ± 4.9 | 77.3 ± 2.4 |
| ResNet-50 | ROC-AUC | 69.1 ± 2.1 | 67.5 ± 5.6 | 72.0 ± 3.9 | 74.6 ± 3.1 | 72.9 ± 6.0 | 75.1 ± 6.1 |
| | AU-PRC | 84.7 ± 1.2 | 84.1 ± 3.4 | 86.9 ± 2.3 | 88.2 ± 1.6 | 87.4 ± 3.2 | 88.2 ± 3.4 |
| | F1 | 68.3 ± 4.3 | 67.6 ± 7.8 | 69.9 ± 4.6 | 73.8 ± 3.7 | 73.1 ± 6.4 | 73.9 ± 6.2 |
| | Sensitivity | 57.9 ± 6.6 | 57.6 ± 9.5 | 59.2 ± 6.4 | 64.7 ± 5.5 | 64.5 ± 8.7 | 64.8 ± 8.3 |
| | Specificity | 70.5 ± 5.6 | 69.1 ± 7.6 | 74.0 ± 6.3 | 72.6 ± 4.1 | 70.0 ± 4.7 | 73.4 ± 7.3 |
| EfficientNet-B0 | ROC-AUC | 67.8 ± 10.4 | 73.0 ± 6.0 | 73.9 ± 4.2 | 73.0 ± 4.3 | 72.0 ± 4.1 | 76.5 ± 3.4 |
| | AU-PRC | 84.1 ± 5.9 | 86.6 ± 3.4 | 87.1 ± 2.6 | 86.8 ± 2.6 | 86.1 ± 2.3 | 88.5 ± 1.9 |
| | F1 | 64.5 ± 22.9 | 73.5 ± 7.1 | 74.7 ± 3.1 | 74.0 ± 3.8 | 73.2 ± 5.5 | 77.4 ± 2.3 |
| | Sensitivity | 56.6 ± 21.3 | 65.2 ± 9.4 | 66.4 ± 5.0 | 65.7 ± 5.2 | 65.0 ± 7.9 | 70.2 ± 3.3 |
| | Specificity | 71.1 ± 12.6 | 70.0 ± 6.0 | 70.1 ± 7.8 | 69.1 ± 5.4 | 68.2 ± 6.0 | 70.9 ± 4.6 |
| EfficientNet-B1 | ROC-AUC | 72.0 ± 4.2 | 71.7 ± 5.4 | 73.5 ± 3.6 | 74.3 ± 2.8 | 74.3 ± 3.2 | 76.0 ± 5.5 |
| | AU-PRC | 86.9 ± 2.5 | 86.1 ± 3.3 | 87.2 ± 2.1 | 87.5 ± 1.6 | 87.6 ± 1.8 | 88.6 ± 3.4 |

| Model | Metric | Baseline | Single-ET | 3Res-ET | Mix5-ET | 5Rad-ET | All8-ET |
|---|---|---|---|---|---|---|---|
| | F1 | 70.6 ± 6.7 | 72.5 ± 4.5 | 74.3 ± 3.1 | 74.7 ± 2.4 | 74.5 ± 3.3 | 76.5 ± 2.9 |
| | Sensitivity | 61.0 ± 9.9 | 63.6 ± 6.4 | 65.8 ± 4.1 | 66.2 ± 4.1 | 65.9 ± 5.0 | 68.8 ± 4.2 |
| | Specificity | 71.7 ± 11.0 | 69.9 ± 7.7 | 70.4 ± 4.7 | 71.2 ± 6.3 | 71.0 ± 4.2 | 71.1 ± 9.0 |
| ViT | ROC-AUC | 51.9 ± 4.5 | 52.8 ± 4.0 | 51.4 ± 3.8 | 51.9 ± 3.3 | 52.2 ± 4.1 | 51.4 ± 3.4 |
| | AU-PRC | 74.5 ± 2.6 | 75.3 ± 2.5 | 74.4 ± 2.0 | 74.5 ± 1.8 | 74.9 ± 2.3 | 74.4 ± 2.0 |
| | F1 | 34.1 ± 23.6 | 54.1 ± 31.7 | 60.1 ± 35.4 | 74.7 ± 25.8 | 59.5 ± 30.0 | 66.2 ± 27.4 |
| | Sensitivity | 25.5 ± 20.3 | 56.8 ± 41.6 | 67.1 ± 43.7 | 87.0 ± 31.8 | 61.5 ± 38.7 | 72.0 ± 38.2 |
| | Specificity | 79.5 ± 16.6 | 45.8 ± 43.1 | 34.8 ± 44.5 | 14.1 ± 32.6 | 41.7 ± 40.3 | 29.7 ± 39.7 |
| Swin | ROC-AUC | 76.8 ± 3.5 | 70.4 ± 10.6 | 67.5 ± 18.3 | 65.7 ± 15.0 | 65.6 ± 17.7 | 70.6 ± 17.0 |
| | AU-PRC | 89.3 ± 1.9 | 85.4 ± 6.3 | 83.6 ± 10.6 | 82.3 ± 9.1 | 82.5 ± 10.5 | 85.2 ± 10.0 |
| | F1 | 75.0 ± 2.9 | 67.8 ± 18.0 | 70.5 ± 23.2 | 72.5 ± 13.2 | 65.0 ± 31.0 | 66.3 ± 33.5 |
| | Sensitivity | 65.5 ± 3.9 | 59.8 ± 19.8 | 67.6 ± 29.4 | 68.2 ± 21.4 | 64.3 ± 36.2 | 64.6 ± 34.6 |
| | Specificity | 76.0 ± 4.7 | 71.5 ± 19.2 | 62.2 ± 33.7 | 57.5 ± 32.7 | 60.2 ± 42.0 | 67.7 ± 32.9 |

**Supplementary Table 7 Per-architecture external validation performance on MIMIC-Eye (N = 1,938, binary: normal vs. abnormal). For each backbone architecture and metric, we report mean ± SD (%) across 10 random seeds. All metrics except ROC-AUC and AU-PRC are computed at the Youden's J-optimal threshold. Single-ET averages across all eight individual-reader strategies. Models were evaluated by inference only using checkpoints trained on the Emory cohort (no retraining).**

| Model | Metric | Baseline | Single-ET | 3Res-ET | Mix5-ET | 5Rad-ET | All8-ET |
|---|---|---|---|---|---|---|---|
| **DenseNet-121** | ROC-AUC | 68.9 ± 4.0 | 67.8 ± 4.9 | 72.1 ± 5.4 | 72.7 ± 5.9 | 71.6 ± 5.5 | 71.8 ± 2.9 |
| | AU-PRC | 58.9 ± 4.1 | 58.8 ± 5.2 | 64.6 ± 6.4 | 65.2 ± 7.2 | 63.0 ± 7.4 | 63.0 ± 4.4 |
| | F1 | 57.2 ± 6.1 | 57.2 ± 5.3 | 61.0 ± 5.3 | 61.4 ± 7.1 | 61.8 ± 3.5 | 61.6 ± 2.3 |
| | Sensitivity | 57.4 ± 9.6 | 59.2 ± 8.7 | 62.5 ± 7.3 | 63.6 ± 11.8 | 68.9 ± 7.3 | 65.0 ± 5.0 |
| | Specificity | 71.6 ± 5.1 | 68.3 ± 7.3 | 71.6 ± 5.7 | 71.7 ± 7.5 | 63.6 ± 10.4 | 68.9 ± 6.9 |
| EfficientNet-B0 | ROC-AUC | 66.9 ± 6.7 | 70.2 ± 4.5 | 71.3 ± 2.9 | 70.2 ± 3.0 | 69.0 ± 3.4 | 70.9 ± 2.9 |
| | AU-PRC | 57.1 ± 6.5 | 59.6 ± 5.0 | 60.9 ± 3.8 | 60.6 ± 3.3 | 58.8 ± 4.0 | 60.8 ± 3.5 |
| | F1 | 52.5 ± 17.7 | 60.4 ± 4.1 | 61.1 ± 3.2 | 59.6 ± 3.4 | 59.0 ± 2.9 | 61.2 ± 3.1 |
| | Sensitivity | 53.9 ± 19.2 | 65.5 ± 6.5 | 65.9 ± 7.0 | 61.8 ± 5.3 | 62.4 ± 6.0 | 67.4 ± 7.0 |
| | Specificity | 72.1 ± 10.8 | 65.5 ± 7.1 | 66.6 ± 5.1 | 69.3 ± 3.4 | 67.0 ± 7.6 | 64.7 ± 5.6 |
| **EfficientNet-B1** | ROC-AUC | 71.2 ± 2.7 | 68.4 ± 4.9 | 70.1 ± 3.2 | 68.8 ± 2.9 | 68.2 ± 3.8 | 70.2 ± 2.8 |
| | AU-PRC | 61.4 ± 2.5 | 57.6 ± 5.3 | 60.2 ± 3.4 | 58.8 ± 3.2 | 58.7 ± 3.5 | 61.4 ± 2.7 |
| | F1 | 59.9 ± 5.2 | 59.8 ± 3.8 | 60.3 ± 3.2 | 58.7 ± 2.6 | 58.3 ± 3.4 | 58.8 ± 3.5 |
| | Sensitivity | 62.5 ± 11.5 | 65.9 ± 7.6 | 64.5 ± 6.4 | 61.3 ± 5.3 | 60.4 ± 6.2 | 59.9 ± 6.9 |
| | Specificity | 70.0 ± 7.5 | 63.2 ± 10.4 | 66.9 ± 6.7 | 68.1 ± 6.8 | 68.4 ± 7.7 | 70.9 ± 4.6 |
| **ResNet-18** | ROC-AUC | 69.9 ± 2.1 | 68.4 ± 5.6 | 71.8 ± 2.4 | 71.6 ± 3.0 | 74.4 ± 2.0 | 75.4 ± 1.1 |
| | AU-PRC | 59.7 ± 2.5 | 60.6 ± 6.4 | 64.7 ± 2.6 | 63.7 ± 3.5 | 67.3 ± 2.3 | 67.9 ± 1.3 |
| | F1 | 60.3 ± 2.4 | 57.2 ± 6.7 | 60.6 ± 2.4 | 61.3 ± 2.6 | 63.2 ± 1.9 | 64.4 ± 1.7 |
| | Sensitivity | 65.7 ± 5.6 | 58.4 ± 10.8 | 61.5 ± 4.4 | 65.8 ± 5.1 | 65.3 ± 4.2 | 67.3 ± 6.1 |
| | Specificity | 64.9 ± 4.9 | 70.1 ± 8.9 | 72.1 ± 3.3 | 67.1 ± 6.2 | 72.2 ± 3.5 | 72.0 ± 5.8 |
| **ResNet-50** | ROC-AUC | 71.4 ± 1.9 | 67.6 ± 4.5 | 67.4 ± 3.2 | 71.3 ± 2.5 | 70.4 ± 6.4 | 72.1 ± 3.6 |
| | AU-PRC | 62.9 ± 2.1 | 59.0 ± 5.1 | 60.4 ± 3.7 | 63.6 ± 3.8 | 63.2 ± 6.7 | 64.6 ± 3.9 |
| | F1 | 60.4 ± 2.1 | 57.2 ± 4.6 | 55.6 ± 2.7 | 59.5 ± 2.8 | 59.3 ± 6.4 | 60.7 ± 3.4 |
| | Sensitivity | 61.3 ± 4.4 | 59.5 ± 9.5 | 54.3 ± 6.2 | 59.6 ± 7.3 | 60.0 ± 8.7 | 62.2 ± 6.7 |
| | Specificity | 71.8 ± 3.8 | 67.6 ± 11.1 | 72.5 ± 7.3 | 72.8 ± 8.7 | 71.7 ± 5.8 | 71.4 ± 6.9 |
| **Swin** | ROC-AUC | 70.7 ± 2.8 | 66.1 ± 8.8 | 65.6 ± 12.8 | 61.7 ± 15.1 | 59.6 ± 17.4 | 66.5 ± 14.1 |
| | AU-PRC | 62.1 ± 2.8 | 57.6 ± 9.8 | 57.7 ± 13.2 | 53.9 ± 15.9 | 52.6 ± 17.5 | 58.8 ± 15.2 |
| | F1 | 60.0 ± 2.7 | 54.9 ± 12.8 | 51.9 ± 21.2 | 58.3 ± 8.8 | 55.4 ± 19.2 | 55.8 ± 18.2 |
| | Sensitivity | 60.6 ± 4.0 | 59.4 ± 18.3 | 58.0 ± 28.9 | 74.9 ± 21.7 | 75.0 ± 30.4 | 64.4 ± 28.1 |
| | Specificity | 72.1 ± 4.9 | 65.0 ± 19.1 | 66.3 ± 28.0 | 44.5 ± 34.6 | 43.9 ± 40.3 | 62.0 ± 31.7 |
| **ViT** | ROC-AUC | 53.1 ± 3.6 | 53.7 ± 4.1 | 50.6 ± 2.9 | 51.9 ± 2.7 | 52.3 ± 2.2 | 50.5 ± 4.1 |
| | AU-PRC | 43.9 ± 3.6 | 44.2 ± 3.7 | 41.7 ± 2.2 | 42.6 ± 2.3 | 43.3 ± 2.0 | 42.2 ± 3.5 |
| | F1 | 48.3 ± 8.0 | 50.0 ± 15.1 | 46.4 ± 20.2 | 51.4 ± 16.9 | 53.4 ± 9.0 | 53.0 ± 12.7 |
| | Sensitivity | 60.4 ± 23.4 | 80.1 ± 34.0 | 74.2 ± 38.2 | 86.2 ± 31.2 | 86.5 ± 28.6 | 88.5 ± 28.6 |
| | Specificity | 45.2 ± 25.9 | 21.6 ± 36.2 | 26.2 ± 38.5 | 14.5 ± 31.6 | 14.5 ± 30.5 | 12.0 ± 29.6 |

**Supplementary Table 8 Per-reader significance for Single-ET training. Each of the eight experts was tested individually against the per-model baseline over n = 70 paired architecture and seed observations, with Bonferroni-adjusted p values (p_adj) over the same 35 strategy-metric tests as Table 2. No reader separates on macro F1. Two readers do separate on the area under the receiver-operating-characteristic curve, which is why the manuscript scopes the claim that no single reader separated to macro F1.**

| Reader | n paired | F1 (pp) | F1 p_adj | ROC-AUC (pp) | ROC-AUC p_adj | AU-PRC (pp) | AU-PRC p_adj |
|---|---|---|---|---|---|---|---|
| Rad1 | 70 | +1.78 | 1.0000 | +1.42 | 1.0000 | +0.69 | 1.0000 |
| Rad2 | 70 | +1.07 | 1.0000 | +0.34 | 1.0000 | -0.06 | 1.0000 |
| Rad3 | 70 | +1.41 | 1.0000 | +1.34 | 1.0000 | +0.80 | 1.0000 |
| Rad4 | 70 | +1.10 | 1.0000 | +0.91 | 1.0000 | +0.28 | 1.0000 |
| Rad5 | 70 | +3.81 | 0.1723 | +3.13 | 0.0077 | +2.72 | 0.1307 |
| Res1 | 70 | +2.10 | 1.0000 | +1.91 | 1.0000 | +1.65 | 1.0000 |
| Res2 | 70 | +2.55 | 0.8684 | +2.44 | 0.0378 | +1.61 | 0.5794 |
| Res3 | 70 | +1.15 | 1.0000 | +1.33 | 1.0000 | +0.81 | 1.0000 |

**Supplementary Table 9 Sensitivity of the mask-coverage analysis to the reading subset and the fixation threshold. Union coverage is the fraction of the image retained by pooling k readers' masks; the pairwise Jaccard index is averaged over reader pairs. The main text quotes the first row, the 3,150 unassisted readings at threshold 0, because an unassisted reading is the correct estimand for what one reader's mask retains. The paper's claim is that coverage front-loads less of its growth into the first three readers than performance does, 71.1%. That holds in all four strata, at 54.7%, 47.4%, 54.8% and 54.3%, so the saturation is not explained by coverage growth alone under any of these choices.**

| Stratum | Definition | Coverage, k = 1 (%) | Coverage, k = 3 (%) | Coverage, k = 8 (%) | Mean pairwise Jaccard | Gain, k = 1 to 3 (pp) | Gain, k = 3 to 8 (pp) |
|---|---|---|---|---|---|---|---|
| Unassisted | Unassisted readings only (3,150 masks), per-reading estimand, threshold 0 | 20.6 | 34.0 | 45.1 | 0.411 | 13.4 | 11.1 |
| AI-assisted | AI-assisted readings only (2,535 masks), per-reading estimand, threshold 0 | 10.3 | 21.1 | 33.2 | 0.225 | 10.8 | 12.0 |
| Unassisted, union | Unassisted readings, union over sessions estimand, threshold 0 | 33.0 | 47.4 | 59.3 | 0.564 | 14.4 | 11.9 |
| Unassisted, threshold 5 | Unassisted readings, per reading, threshold 5 | 17.8 | 30.5 | 41.3 | 0.370 | 12.8 | 10.7 |

**Supplementary Table 10 Per-reader external transfer. Each expert's Single-ET models were compared with the per-model baseline on each external cohort over n = 70 paired architecture and seed observations, using the learning-rate-matched checkpoint set. Deltas are mean percentage-point differences from the baseline and p values are two-sided paired Wilcoxon signed-rank tests, uncorrected. Rad5 has the largest in-domain macro F1 gain of the eight readers (Supplementary Table 8) yet the lowest ROC-AUC delta on both cohorts, Res2 is the only reader whose difference is positive and reaches $p < 0.05$ on either cohort; Rad4 and Rad5 reach it on MIMIC-Eye as declines.**

| Reader | n paired | SIIM ROC-AUC (pp) | SIIM p | SIIM F1 (pp) | MIMIC ROC-AUC (pp) | MIMIC p | MIMIC F1 (pp) |
|---|---|---|---|---|---|---|---|
| Rad1 | 70 | +0.08 | 0.9510 | +4.34 | -1.52 | 0.0983 | -1.47 |
| Rad2 | 70 | +0.54 | 0.5604 | +2.91 | -1.03 | 0.1663 | +0.50 |
| Rad3 | 70 | +0.50 | 0.6756 | +5.35 | -0.95 | 0.2674 | +0.57 |
| Rad4 | 70 | -0.04 | 0.7542 | +3.54 | -1.92 | 0.0108 | -1.63 |
| Rad5 | 70 | -1.55 | 0.3387 | +2.35 | -2.37 | 0.0203 | -0.47 |
| Res1 | 70 | -0.69 | 0.5293 | +3.64 | -1.44 | 0.1443 | +0.09 |
| Res2 | 70 | +2.02 | 0.0136 | +4.98 | -0.13 | 0.9091 | +0.04 |
| Res3 | 70 | -0.62 | 0.8768 | +1.99 | -1.84 | 0.0692 | +0.08 |